\documentclass[10pt,twocolumn,letterpaper]{article}

\usepackage{cvpr}
\usepackage{times}
\usepackage{epsfig}
\usepackage{graphicx}
\usepackage{amsmath}
\usepackage{amssymb}

\usepackage{multirow}
\newcommand{\tabincell}[2]{\begin{tabular}{@{}#1@{}}#2\end{tabular}} 
\usepackage{makecell}
\usepackage{hhline}
\usepackage{enumitem}
\setenumerate[1]{itemsep=0pt,partopsep=0pt,parsep=\parskip,topsep=5pt}
\setitemize[1]{itemsep=0pt,partopsep=0pt,parsep=\parskip,topsep=5pt}
\setdescription{itemsep=0pt,partopsep=0pt,parsep=\parskip,topsep=5pt}
\usepackage{comment}
\usepackage[export]{adjustbox}
\usepackage{authblk}

\cvprfinalcopy 

\def\cvprPaperID{3096} 

\ifcvprfinal \fi
\begin{document}

\title{Temporal Attention-Gated Model for Robust Sequence Classification}

\author{Wenjie Pei\textsuperscript{1}, Tadas Baltru\v{s}aitis\textsuperscript{2}, David M.J. Tax\textsuperscript{1} and Louis-Philippe Morency\textsuperscript{2}\\
\vspace{-3mm}
\textsuperscript{1}Pattern Recognition Laboratory, Delft University of Technology\\
\textsuperscript{2}Language Technologies Institute, Carnegie Mellon University\\
{\tt\small W.Pei-1@tudelft.nl, tbaltrus@cs.cmu.edu, D.M.J.Tax@tudelft.nl, morency@cs.cmu.edu}
}

\maketitle
\thispagestyle{empty}

\begin{abstract}

Typical techniques for sequence classification are designed for well-segmented sequences which have been edited to remove noisy or irrelevant parts. Therefore, such methods cannot be easily applied on noisy sequences expected in real-world applications. In this paper,  we present the Temporal Attention-Gated Model (TAGM) which integrates ideas from attention models and gated recurrent networks to better deal with noisy or unsegmented sequences. Specifically, we extend the concept of attention model to measure the relevance of each observation (time step) of a sequence. We then use a novel gated recurrent network to learn the hidden representation for the final prediction. An important advantage of our approach is interpretability since the temporal attention weights provide a meaningful value for the salience of each time step in the sequence. We demonstrate the merits of our TAGM approach, both for prediction accuracy and interpretability, on three different tasks: spoken digit recognition, text-based sentiment analysis and visual event recognition.
\end{abstract}
\vspace{-5mm}

\section{Introduction}
\vspace{-1mm}
Sequence classification is posed as a problem of assigning a label to a sequence of observations. Sequence classification models have extensive applications ranging from computer vision~\cite{video_classification} to natural language processing~\cite{Bahdanau14}. Most existing sequence classification models are designed for well segmented sequences and do not explicitly model the fact that  irrelevant (noisy) parts may be present in the sequence.  To reduce the interference of these irrelevant parts, researchers will often manually pre-process the dataset to remove irrelevant subsequences. This manual pre-processing can be very time consuming and reduce applicability in real-world scenarios. 

A popular approach for sequence classification is gated recurrent networks like Gated Recurrent Units (GRU)~\cite{GRU} and Long Short-Term Memory (LSTM)~\cite{LSTM}. They employ gates (e.g., the input gate in the LSTM model) to balance between current and previous time steps when memorizing the temporal information flow. However, these vectorial gates are applied individually to each dimension of the information flow, thus it is hard to interpret the relative importance of the input time observations (i.e., time steps). What subset of sequential observations is the most salient for the classification task? Another way to balance the information flow, as we do in this work,  is the adoption of attention-based mechanism, which applies individual attention scores to each observation (time step), allowing for better interpretability.
\begin{figure}[!t]
\centering
   \includegraphics[width=0.9\linewidth]{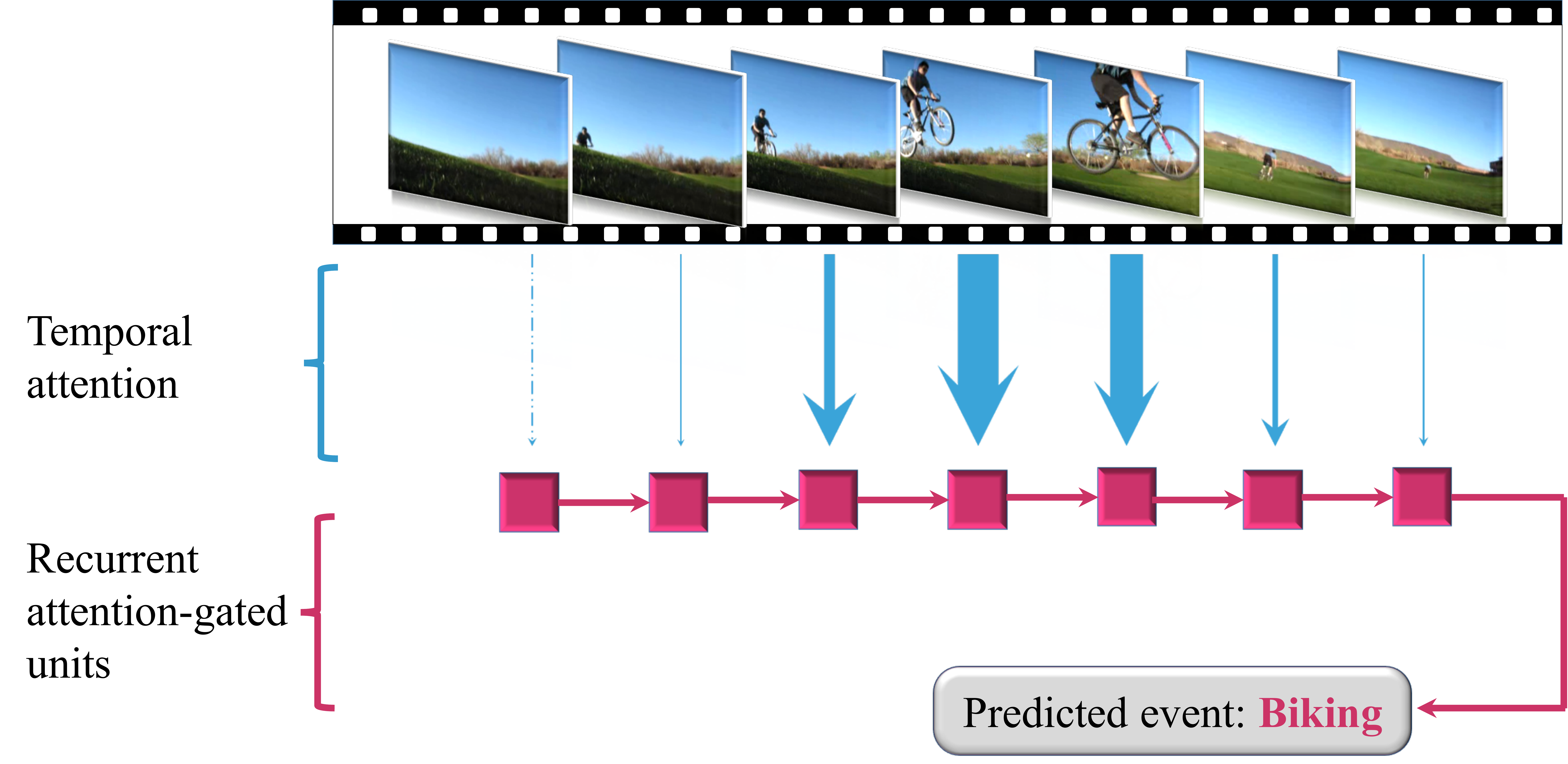} \\
   \caption{Our proposed model first employs an attention module to extract the salient frames from the noisy raw input sequences, and then learns an effective hidden representation for the top classifier. The wider the arrow is, the more the information is incorporated into the hidden representation. The dashed line represents no transfer of information.}
\label{fig:overview}
\vspace{-2mm}
\end{figure}

In this paper, we introduce the Temporal Attention-Gated Model (TAGM) which extends the idea of attention-based mechanism to sequence classification tasks (see overview in Figure~\ref{fig:overview}). TAGM's attention module automatically localizes the salient observations which are relevant to the final decision and ignore the irrelavant (noisy) parts of the input sequence. We created a new recurrent neural unit that can learn a better sequence hidden representation based on the attention scores. Consequently, TAGM's classification decision is made based on the selected relevant segments, improving accuracy over the conventional models that take into account the whole input sequence. 

Notably, compared to conventional sequence classification models, TAGM benefits from the following advantages:
\begin{itemize}
\setlength{\itemsep}{0pt}
\setlength{\parsep}{0pt}
\setlength{\parskip}{0pt}
\item  It is able to automatically capture salient parts of the input sequences thereby leading to better performance. 
\item The inferred attention (scalar) scores provide a meaningful interpretation for the informativeness of each observation in the sequence. 
\item Compared to conventional gated recurrent models such as LSTM, our model reduces the number of parameters which leads to faster training and inference and better generalizability with less training data.
\item The proposed model is able to generalize to tasks in computer vision, speech recognition, and natural language processing.
\end{itemize}
\vspace{-3mm}
\section{Related Work}
While a full review of previous sequence classification models is beyond the scope of this paper, in this section we summarize approaches most relevant to our proposed approach, grouping them in three areas: sequence classification, attention models and recurrent networks. \\
\textbf{Sequence Classification.}
The conventional sequence classification models can be divided roughly into two categories: generative and discriminative models.

The first category focuses on learning an effective intermediate representation based on generative models.
These methods are typically based on the Hidden Markov Models (HMMs)~\cite{HMM}. The HMM is a generative model which can be extended to class-conditional HMMs for sequence classification by combining class priors via Bayes' rule. HMM can also be used as the base model for Fisher Kernel~\cite{fisher_kernel} to learn a sequence representation.

The second category is the discriminative graphical models which model the distribution over all class labels conditioned on the input data. Conditional random fields (CRF)~\cite{CRF} are discriminative models for sequence labeling which aims to assign one label for each sequence observation. A potential drawback of common CRFs is that the linear mapping between observations and labels cannot model complex decision boundaries, which gives rise to many non-linear CRF-variants (e.g., latent-dynamic CRFs~\cite{latentCRF}, conditional neural fields~\cite{CNF}, neural conditional random fields~\cite{Do_AISTATS_2010} and hidden-unit CRF model~\cite{HuCRF}). Hidden-state CRF  (HCRF)~\cite{HCRF} employs a chain of k-nomial latent variables to model the latent structure and has been successfully used in the sequence labeling. Similarly, hidden unit logistic model (HULM)~\cite{HULM} utilizes binary stochastic hidden units to represent the exponential hidden states so as to model more complex latent decision boundaries.  

Aforementioned works are specifically designed for well segmented sequences and hence cannot cope well with noisy or unsegmented sequences.\\
\textbf{Attention Models.}
Inspired by the attention scheme of human foveal vision, 	attention model was proposed to focus selectively on certain relevant parts of the input by measuring the sensitivity of output to variances of the input. Doing so can not only improve the performance of the model but can also result in better interpretability~\cite{Kelvin2015}. Attention models have been applied to image and video captioning~\cite{Kelvin2015, ChenZ15, Fang15, yao2015}, machine translation~\cite{Bahdanau14, luong15attn, SankaranMAI16}, depth-based person identification~\cite{Haque_2016_CVPR} and speech recognition~\cite{Graves06}. To the best of our knowledge, our TAGM is the first
end-to-end recurrent neural network to employ the attention mechanism in the temporal domain of sequences, with the added advantage of interpretability of its temporal salience indicators (i.e., temporal attention) at each time step (sequence observation). Our work is different from prior work focused on spatial domain (e.g., images) such as the model proposed by Sharma et al.~\cite{SharmaKS15}. \\
\textbf{Recurrent Networks.}
Recurrent Neural Networks (RNN) learn a representation for each time step by taking into account both the observation at current time step and the representation in the previous one~\cite{Schmidhuber1989}. The biggest advantage of recurrent neural networks lies in their capability of preserving information over time by the recurrent mechanism. Recurrent networks have been successfully applied to various tasks including language modeling~\cite{Mikolov11}, image generation~\cite{Theis_2015} and online handwriting generation~\cite{Graves13}. To address the gradient vanishing problem of plain-RNN when dealing with long sequences, LSTM~\cite{LSTM} and GRU~\cite{GRU} were proposed. They are equipped with the gates to balance the information flow from the previous time step and current time step dynamically. Inspired by this setup, our TAGM model also employs a gate to filter out the noisy time steps and preserve the salient ones. The difference from the LSTM and GRU is that the gate value in our model is fed from the attention module which focuses on learning the salience at each time step.

\vspace{-2mm}
\section{Temporal Attention-Gated Model}

\begin{figure}[!htb]
\centering
   \includegraphics[width=0.8\linewidth]{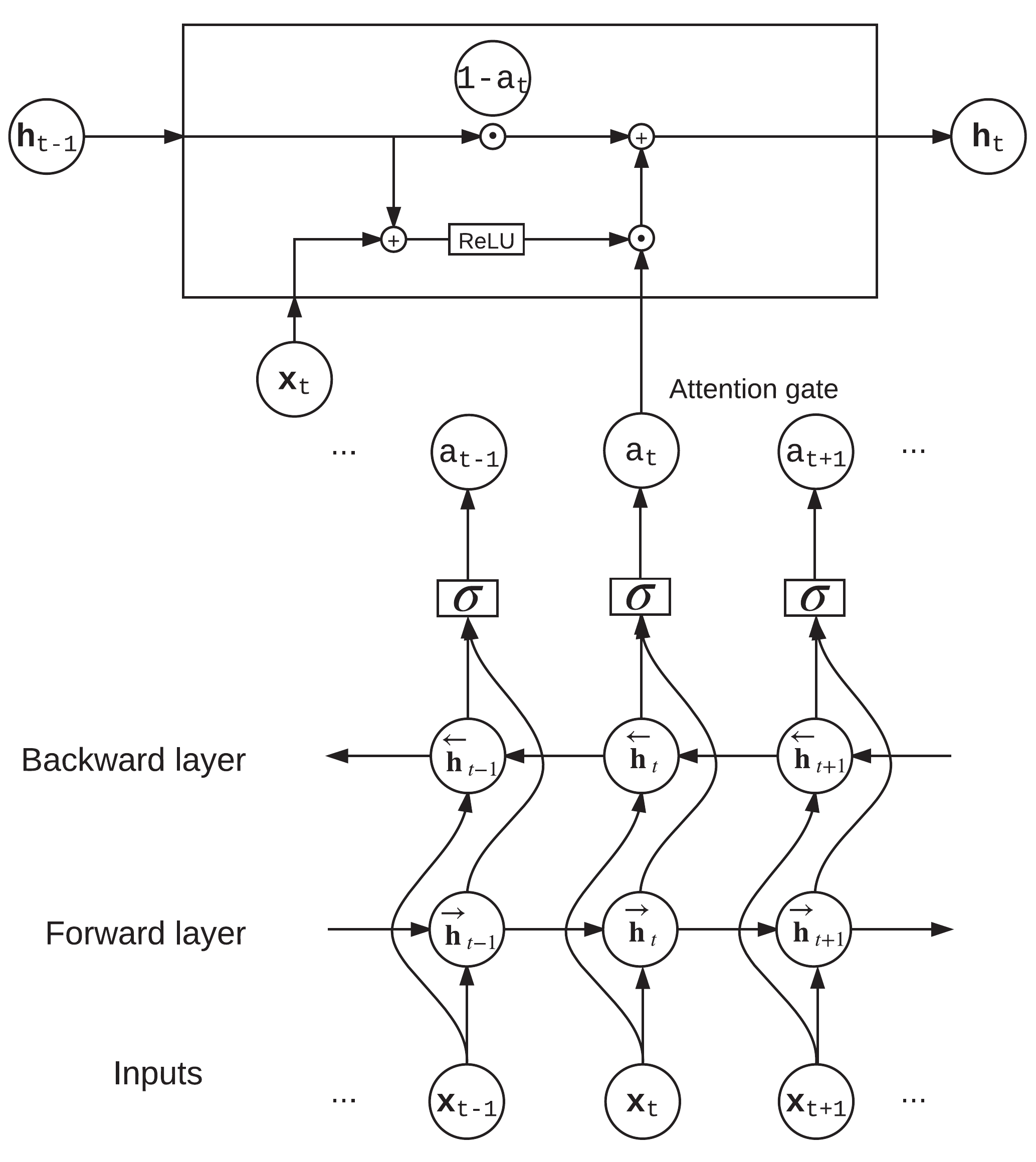} \\
\vspace{1mm}
   \caption{ The graphical representation of our Temporal Attention-Gated Model (TAGM). The top part of the figure is the Recurrent Attention-Gated Units and the bottom is the Temporal Attention Module. Note that $a_t$ is the saliency score represented as a  scalar value instead of a vector, hence $\odot$ in the figure means multiplication between a scalar and a vector.}
\label{fig:model}
\end{figure}

Given as input an unsegmented sequence of possibly noisy observations, our goal is to: (1) calculate a salience score for each time step observation in our input sequence, and (2) construct a hidden representation based on the salience scores, best suited for the sequence classification task. To achieve these goals, we propose the Temporal Attention-Gated Model (TAGM) which consists of two modules: temporal attention module, and recurrent attention-gated units. Our TAGM model can be trained in an end-to-end manner efficiently. The graphical structure of the model is illustrated in Figure~\ref{fig:model}.  

\vspace{-2mm}
\subsection{Recurrent Attention-Gated Units}
\vspace{-1mm}
The goal of the recurrent attention-gated units is to learn a hidden sequence representation which integrates the attention scores (inferred from the temporal attention module that will be discussed in the next section). In order to integrate the attention scores in the recurrent network units, we define an attention gate to control how much information is incorporated from the input of the current time step based on the salience and relevance to the final task. 

Formally, given an input sequence $\mathbf{x}_{1, \dots, T} = \{ \mathbf{x}_1, \dots, \mathbf{x}_T \}$ of length $T$ in which $\mathbf{x}_t \in \mathbb{R}^D$ denotes the observation at the $t$-th time step, the attention score at time step $t$ is denoted as $a_t$, which is a scalar value that indicates the salience of current time step to the final decision.  For this purpose, we define our core recurring process where the hidden state $\mathbf{h}_t$ at time step $t$ is modeled as a convex summation:
\begin{equation}
\mathbf{h}_{t} = (1-a_t) \cdot \mathbf{h}_{t-1} + a_t \cdot \mathbf{h'}_{t} 
\label{eqn:hidden_repre}
\end{equation}
Wherein, $\mathbf{h}_{t-1}$ is the previous hidden state and $\mathbf{h'}_{t}$ is the candidate hidden state value which fully incorporates the input information $\mathbf{x}_t$ in the current time step:
\begin{equation}
\mathbf{h'}_{t} = g(\mathbf{W} \cdot \mathbf{h}_{t-1} + \mathbf{U} \cdot \mathbf{x}_t + \mathbf{b} )
\end{equation}
Herein, $\mathbf{W}$ and $\mathbf{U}$ are respectively the linear transformation parameters for previous and current time steps while $\mathbf{b}$ is the bias term. We use the rectified linear unit (ReLU)\cite{ReLU} as the activation function $g$. Equation~\ref{eqn:hidden_repre} uses attention score $a_t$ to balance the information flow between current candidate hidden state $\mathbf{h'}_{t}$ and previous hidden state $\mathbf{h}_{t-1}$. High attention value will push the model to focus more on the current hidden state $\mathbf{h'}_{t}$ and input feature $\mathbf{x}_t$, while low attention value would make the model ignore the current input feature and inherit more information from previous time steps.

The learned hidden representation at the last time step $\mathbf{h}_T$ of the sequence is further fed into the final classifier, often a softmax function, to perform a classification task, which calculates the probability of a predicted label $y_k$ among $K$ classes as: 
{\setlength{\abovedisplayskip}{4pt}
\setlength{\belowdisplayskip}{2pt}
\begin{equation}
\begin{small}
P(y_k | \mathbf{h}_{T}) = \frac{\mathrm{exp}\{\mathbf{W}^\top_k \mathbf{h}_T + b_k\}}{\sum_{i=1}^K\mathrm{exp}\{ \mathbf{W}^\top_i \mathbf{h}_T + b_i \}}
\label{eqn:softmax}
\end{small}
\end{equation}}\\
where $\mathbf{W}^\top_i$ and $b_i$ refer to the parameters calculating the linear mapping score for the $i$-th class.
\vspace{-2mm}
\subsection{Temporal Attention Module}
\vspace{-1mm}
\label{sec:attention}
The goal of this module is to estimate the saliency and relevance of each sequence observation. This saliency score should not only be based on the input observation at the current time step, but also take into consideration information from neighboring observations in both directions. To model this neighborhood influence, we infer the attention score $a_t$ in Equation~\ref{eqn:hidden_repre} using a bi-directional RNN:
{\setlength{\abovedisplayskip}{4pt}
\setlength{\belowdisplayskip}{4pt}
\begin{equation}
a_t = \sigma(\mathbf{m}^\top (\overrightarrow{h}_t;  \overleftarrow{h}_t)+b )
\label{eqn:attention}
\end{equation}
}Herein, $\mathbf{m}$ is the weight vector of our fusion layer which integrates both directional layers of our bi-directional RNN and $b$ is the bias term. A sigmoid function is employed as the activation function $\sigma$ at the top layer of the attention module in Equation~\ref{eqn:attention} to constraint the attention weight to lie between $[0, 1]$.  $\overrightarrow{h}_t$ and $\overleftarrow{h}_t$ are the hidden representations of a bi-directional RNN model:
{\setlength{\abovedisplayskip}{4pt}
\setlength{\belowdisplayskip}{4pt}
\begin{equation}
\overrightarrow{h}_t = g(\overrightarrow{\mathbf{W}} \mathbf{x}_t+\overrightarrow{\mathbf{U}} \mathbf{\overrightarrow{h}_{t-1} + \overrightarrow{\mathbf{b}}})
\end{equation}
\begin{equation}
\overleftarrow{h}_t = g(\overleftarrow{\mathbf{W}} \mathbf{x}_t+\overleftarrow{\mathbf{U}} \mathbf{\overleftarrow{h}_{t+1} + \overleftarrow{\mathbf{b}}})
\end{equation}
}The ReLU functions are used as the activation functions $g$. Our choice of using plain bi-directional RNN model is motivated by the 
design goal of reducing the number of parameters in our model. 

The learned attention weights $a_t$ serve as the attention gate for Recurrent Attention-Gated Units to control the involved information flow. Furthermore, another important role the learned attention weights play is to provide an interpretability about the degree of salience of each time step.
\vspace{-2mm}
\subsection{End-to-End Parameter Learning}
\vspace{-1mm}
Suppose we are given a training set $\mathcal{D} = \{ (\mathbf{x}^{(n)}_{1, \dots, T}, y^{(n)} ) \}_{n = 1, \dots, N}$ containing $N$ sequences  of length $T$ and their associated labels $y^{(n)}$. $\mathbf{x}^{(n)}_t \in \mathbb{R}^D$ denotes the observation at the $t$-th time step of the $n$-th sample and $T$ can differ from sequence to sequence. We learn jointly the two TAGM modules (temporal attention module and recurrent attention-gated units) and the final sequence classifier by minimizing the conditional negative log-likelihood of the training data with respect to the parameters:
{\setlength{\abovedisplayskip}{1pt plus 3pt minus 2pt}
\setlength{\belowdisplayskip}{1pt plus 3pt minus 2pt}
\begin{align}
\mathcal{L} = -\sum_{n=1}^N \log P \left(\mathbf{y}^{(n)} | \mathbf{x}^{(n)}_{1,\dots, T}\right)
\label{eqn:L}
\end{align}
}\\
Since all three modules (including the final sequence classifier) are analytically differentiable, our TAGM model can be readily trained in an end-to-end manner. The loss is back-propagated through top recurrent attention-gated units and temporal attention module successively using back-propagation through time algorithm~\cite{BPTT}.  
\vspace{-2mm}
\subsection{Comparison with LSTM and GRU}
\vspace{-1mm}
While our model is similar to RNN variants like GRU and LSTM, it is specifically designed with salience detection in mind and has four key differences when compared to them:
\begin{itemize}
\setlength{\itemsep}{0pt}
\setlength{\parsep}{0pt}
\setlength{\parskip}{0pt}
\item We only focus on one scalar attention score to measure the relevance of the current time step instead of generally modeling gate's multi-dimensional values for each hidden unit as done by GRU and LSTM. In this way,  we can obtain an interpretable salience detection (demonstrated on three tasks in Section~\ref{sec:experiment}).
\item We separate the attention modeling and recurrent hidden representation learning as two independent modules to decrease the degree of coupling. One of the advantages of this is our ability to customize the specific recurrent structure for each module with different complexity according to the requirements (eg., different size of hidden units in two modules of TAGM in Table~\ref{table:arabic}). 
\item We employ a bi-directional RNN to take into account both the preceding and the following information of the sequence in the temporal attention module. It helps to model the temporal smoothness of the sequence of salience scores (demonstrated in Figure~\ref{fig:attention_sig}). It should be noted that it is different from the design of the gates in the bi-directional LSTM model since the latter just concatenates the hidden representations of two unidirectional LSTMs, which does not remedy the downside that all vectorial gates are still calculated by considering only one-directional information.
\item Our model only contains one scalar gate, namely the attention gate,  rather than 2 vectorial gates in GRU and 3 gates in LSTM. Doing so enforces the attention gate to take full responsibility of modeling all the salience information. In addition, the model contains fewer parameters (compared to LSTM) and simpler gate structure with less redundancy (compared to GRU and LSTM).  It eases the training procedure and can alleviate the potential over-fitting and has better generalization given small amount of training data, which is demonstrated in Section~\ref{sec:comparison}.      
\end{itemize}
\vspace{-3mm}

\section{Experiments}
\label{sec:experiment}
We performed experiments with TAGM on three publicly available datasets , selected to show generalization across different tasks and modalities: (1) speech recognition on an audio dataset, (2) sentiment analysis on a text dataset, and (3) event recognition on a video dataset.

\textbf{Experimental setup shared across experiments.} 
For all the recurrent networks mentioned in this work (TAGM, GRU, LSTM and plain-RNN), the number of hidden units is tuned by selecting the best configuration from the option set $\{64, 128, 256\}$ using a validation set. The dropout value is validated from the option set $\{0.0, 0.25, 0.5\}$ to avoid potential overfitting. We employ RMSprop as the gradient descent optimization algorithm with gradient clipping between $-5$ and $5$ \cite{gradientclip}. 

We validate the learning rate for parameters $\textbf{m}$ and $b$ in Equation~\ref{eqn:attention} to make the effective region of the sigmoid function of TAGM model adaptive to the specific data. Larger learning rate leads to sharper distribution of attention weights. Code reproducing the results of our experiments is available \footnote{https://github.com/wenjiepei/TAGM}.
\vspace{-1mm}
\subsection{Speech Recognition Experiments}
We first conduct preliminary experiments on a modified dataset constructed from the Arabic spoken digit dataset~\cite{Arabic} to (1) evaluate the effectiveness of the two main modules of TAGM; (2) compare the generalizability of three different gate-setup recurrent models (TAGM, GRU and LSTM) with the varying size of the training data.
\vspace{-4mm}
\subsubsection{Dataset}
\vspace{-2mm}
The Arabic spoken digit dataset contains 8800 utterances, which were collected by asking 88 Arabic native speakers to utter all 10 digits ten times. Each sequence consists of 13-dimensional Mel Frequency Cepstral Coefficents (MFCCs) which were sampled at 11,025Hz, 16-bits using a Hamming window.  We append white noise to the beginning and the end of each sample to simulate the problem with unsegmented sequences. The length of the unrelated sub-sequences before and after the original audio clips is randomized to ensure that the model does not learn to just focus on the middle of the sequence. 
\vspace{-4mm}
\subsubsection{Experimental Setup}
\vspace{-2mm}
We use the same data division as Hammami and Bedda~\cite{Arabic}: 6600 samples as training set and 2200 samples as test set.  We further set aside 1100 samples from training set as the validation set. There is no subject overlap in the three sets.

We compare the performance of our TAGM with three types of baseline models: \\
\textbf{Attention Module + Neural Network (AM-NN).}
To study the impact of our recurrent attention-gated unit, we include a baseline model which employs a feed-forward network directly on top of the temporal attention module. In this AM-NN model, $\mathbf{v}$ is defined as the weighted sum of input features:
\vspace{-2mm}
{\setlength{\abovedisplayskip}{6pt}
\setlength{\belowdisplayskip}{4pt}
\begin{align}
& \mathbf{v} = \sum_{t=1}^{T} {a_t \cdot \mathbf{x}_t}, \quad \mathbf{h} = g (\mathbf{W} \cdot \mathbf{v} + \mathbf{b})
\end{align}
}\\
Sequence classification is performed by passing $\mathbf{h}$ into a softmax layer, as done for our TAGM (see Equation~\ref{eqn:softmax}).\\
\textbf{Discriminative Graphical Models.}
 HCRF and HULM are both extensions of CRF~\cite{CRF} by inserting hidden layers to model the non-linear latent structure in the data. The difference lies in the structure of hidden layers: HCRF uses a chain of $k$-nomial latent variables while HULM utilizes $k$ binary stochastic hidden units. \\
\textbf{Recurrent neural networks.}
Since our model is a recurrent network equipped with a gate mechanism, we compare it with other recurrent networks: plain-RNN, GRU, LSTM. We also investigate the bi-directional variant of our TAGM model (referred as Bi-TAGM), which employs the bi-directional recurrent configuration in the recurrent attention-gated units.

In our experiments,  we also evaluate the generalizability when varying size of training data: from 1,100 to 5,500 training samples. During these experiments, the optimal configuration is selected automatically during validation from the option set \{64,128,256\}.
\vspace{-2mm}
\subsubsection{Results and Discussion }
\vspace{-2mm}
\paragraph{Evaluation of Classification Performance}
Table~\ref{table:arabic} presents the classification performance of several sequence classifiers on Arabic dataset. In order to investigate the effect of the manually added noise information, we perform experiments on both clean and noisy versions of data. 

While the Plain-RNN is unable to recognize spoken digits in a noisy setting, other three recurrent models with gate-setup do not suffer from the noise and obtain comparable performance with the result achieved by HCRF on clean data.  Our model achieves the best results among all classifiers with single-directional recurrent configuration. This probably results from better generalization of our model on the relatively small dataset due to the simpler gate setup and also the attention mechanism. We also perform experiments with the bi-directional version of GRU, LSTM and TAGM, in which our Bi-TAGM  performs best. Bi-GRU achieves its best performance with 64 hidden units. It is worth mentioning that our (single-directional) TAGM using 47 K parameters already achieves comparable result with the Bi-LSTM and Bi-GRU, which indicates that the bi-directional mechanism in the attention module of TAGM enables it to capture most bi-directional information in the attention layer alone. 

\begin{figure}[!htb]
\centering
   \includegraphics[width=0.7\linewidth]{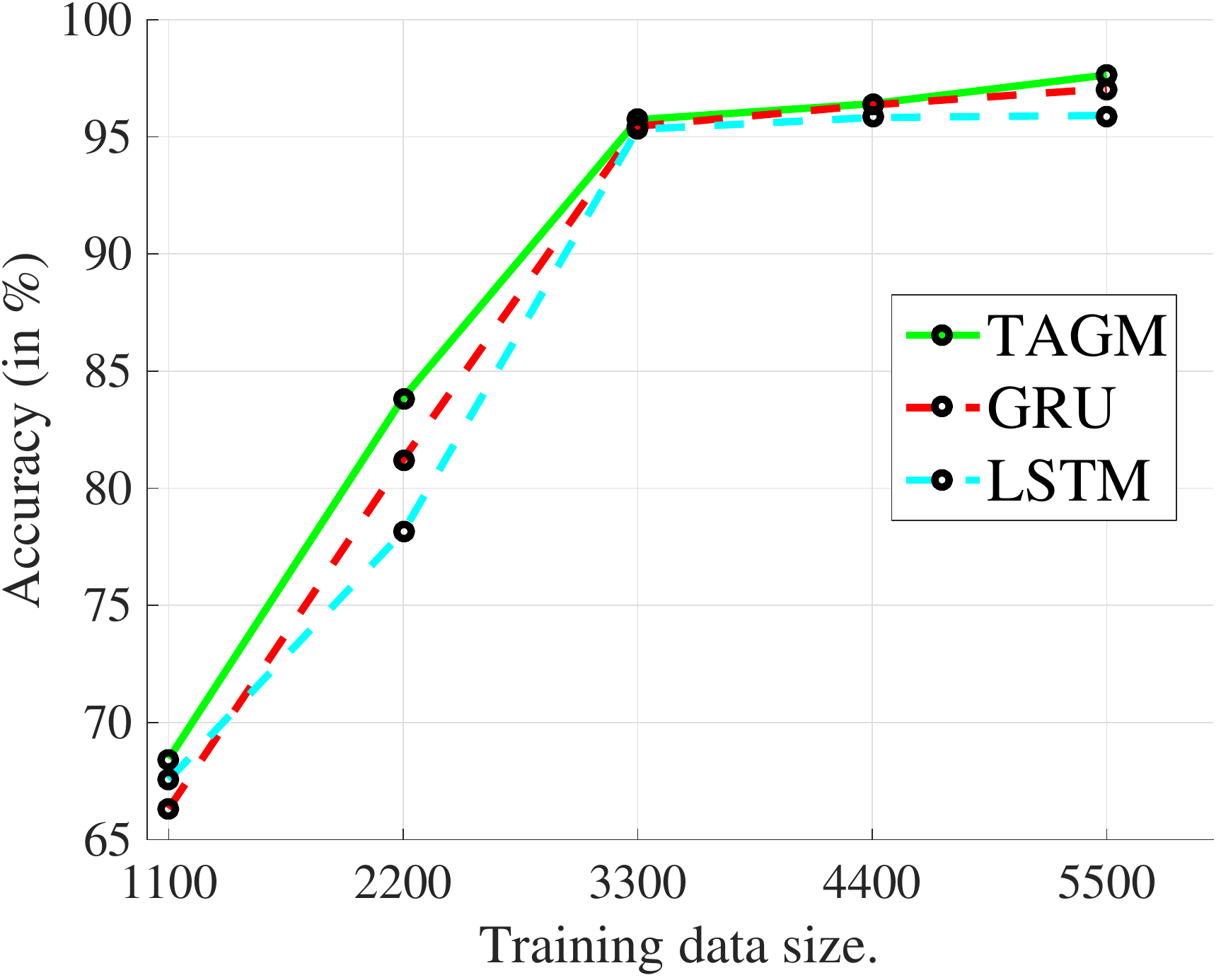} \\
   \caption{ The classification accuracy on the noisy Arabic speech dataset as a function of the size of training data. Note that our TAGM model outperforms GRU and LSTM when less training data is available.}
\label{fig:arabic_comp}
\end{figure}

\begin{figure}[!t]
\begin{minipage}[l]{1.0\linewidth} 
\centering
   \includegraphics[width=0.17\linewidth]{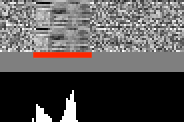} 
   \includegraphics[width=0.17\linewidth]{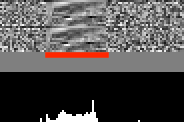} 
   \includegraphics[width=0.17\linewidth]{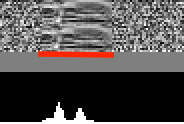} 
   \includegraphics[width=0.17\linewidth]{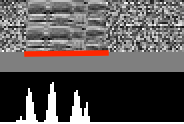} 
   \includegraphics[width=0.17\linewidth]{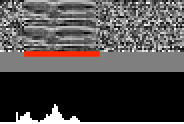} \\ \vspace{1mm}
   \includegraphics[width=0.17\linewidth]{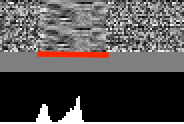} 
   \includegraphics[width=0.17\linewidth]{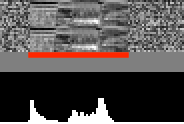} 
   \includegraphics[width=0.17\linewidth]{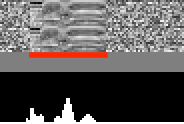} 
   \includegraphics[width=0.17\linewidth]{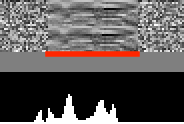} 
   \includegraphics[width=0.17\linewidth]{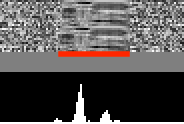}  \\
   (a) \\
   \vspace{1mm}
      \includegraphics[width=0.17\linewidth]{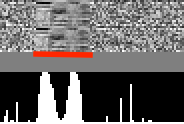} 
   \includegraphics[width=0.17\linewidth]{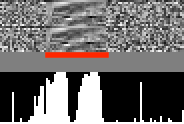} 
   \includegraphics[width=0.17\linewidth]{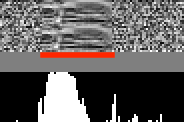} 
   \includegraphics[width=0.17\linewidth]{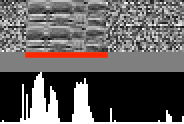} 
   \includegraphics[width=0.17\linewidth]{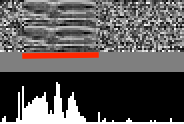} \\ \vspace{1mm}
   \includegraphics[width=0.17\linewidth]{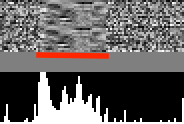} 
   \includegraphics[width=0.17\linewidth]{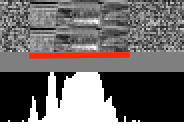} 
   \includegraphics[width=0.17\linewidth]{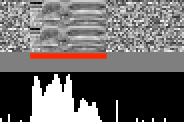} 
   \includegraphics[width=0.17\linewidth]{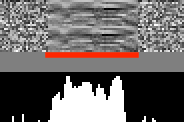} 
   \includegraphics[width=0.17\linewidth]{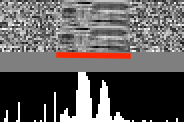} \\
   (b)\\
\end{minipage}
\vspace{1mm}
   \caption{ The visualization of attention weights of TAGM in Figure a and Attention module+NN in Figure b (the weighted features are fed into Feed-forward Neural Networks) on 10 samples (one sample for each digit). For each subfigure, the top subplot shows the spectrogram of the original sequence data, the bottom subplot shows the attention values $a_t$ over time. The red lines indicate the ground-truth of salient segments. Note that TAGM attention weights result in a cleaner attention representation.}
\label{fig:attention_sig}
\end{figure}

\begin{table}[!t]
\caption{Classification accuracy (\%) on  Arabic spoken digit dataset  by
different sequence classification models. Asterisked models ($*$) are trained and evaluated on the clean version of data. Note that we can customize separately the complexity of TAGM's two modules. This design advantage is shown when looking at the optimal TAGM model (after validation) which has 128 dimensions for the Temporal Attention Module, and 64 dimensions for the Recurrent Attention-Gated Units. 
}
\centering
\vspace{2mm}
\normalsize
\resizebox{0.7\linewidth}{!}{
\begin{tabular}{l|c|c|c}
\Xhline{1.0pt}
Model & \#\text{Hidden units} &$\#\text{Parameters}$ & Accuracy \\
\hline
$\text{HULM}^*$~\cite{HULM} & $-$&$-$ & 95.32 \\
$\text{HCRF}^*$~\cite{HULM}&$-$& $-$ & 96.32 \\
HULM & $-$& $-$& 88.27 \\
HCRF &$-$& $-$ &  90.41\\
\hline
$\text{Plain-RNN}^*$&256&75 K & 94.95 \\
Plain-RNN& 256 & 75 K &10.95 \\
GRU & 128 & 61 K & 97.05 \\
LSTM &128  & 81 K  &95.91\\
\hline
NN & 64 & 2.4 K& 65.50  \\
AM-NN & 128-64 &43 K &85.59 \\
\hline
TAGM & 128-64 & 47 K  & \textbf{97.64} \\
\hhline{====}
\Xhline{1.0pt}
Bi-GRU & 64 & 37 K&  97.68\\
Bi-LSTM &256 & 587 K  & 97.45\\
Bi-TAGM & 128-128 & 83 K  & \textbf{97.91} \\
\hline
\end{tabular}}
\label{table:arabic}
\end{table}
\vspace{-3mm}
\paragraph{Comparison of generalizability with the varying size of training data.}
\label{sec:comparison}
We first conduct experiments to compare the generalizability of TAGM to GRU and LSTM by varying the size of training data on the noisy Arabic dataset. Figure~\ref{fig:arabic_comp} presents the experimental results. It can be seen that TAGM exhibits better generalizability than GRU and LSTM on smaller training data sizes, which we believe is caused by the need to learn fewer model parameters, avoiding overfitting.
\vspace{-6mm}
\paragraph{Sequence Salience Detection.}
In order to evaluate the performance of sequence salience detection by our TAGM model, we visualize the attention weights of our model trained on the noisy Arabic dataset, which is illustrated in Figure~\ref{fig:attention_sig}.a.
It shows that the attention model can correctly detect the informative section of the raw signal. 

To investigate  the effect of the temporal information contained in the hidden representation, we also visualize the attention weight of the Attention module + Neural Network classifier, which is shown in Figure~\ref{fig:attention_sig}.b. It shows that the TAGM results in a cleaner and smoother attention weight profile, also notice the spiky behavior, which is mainly achieved by the bi-directional RNN in our temporal attention module.


\vspace{-2mm}
\subsection{Sentiment Analysis Experiments}
Sentiment analysis is a popular research topic in the field of natural language processing (NLP) which aims to identify the viewpoint(s)
underlying a text span~\cite{sentiment_analysis}. We conduct experiments for sentiment analysis to evaluate the performance of our TAGM model on the 	text modality. 
\vspace{-4mm}
\subsubsection{Dataset}
\vspace{-2mm}
The Stanford Sentiment Treebank (SST)~\cite{SST} is a data corpus of movie review excerpts. It consists of 11,855 sentences each of which is assigned a score to indicate the sentimental attitude towards the movie reviews. The dataset offers two types of annotations, sentiment annotations at the sentence level (with a total of 11,855 sentences) and at the phrase level (with a total of 215,154 phrases). The sentence-level and phrase-level labels are provided with two resolutions: binary-classification task (positive or negative) and fine-grained task (5-level classes).
\vspace{-5mm}
\subsubsection{Experimental Setup}
\vspace{-2mm}
Following previous work~\cite{SST}, we utilize 300-d \emph{Glove} word vectors (300 dimensions) pretrained over the Common Crawl~\cite{glove} as the features for each word of the sentences. Our model is well suited to perform sentiment analysis using sentence-level labels. Nevertheless, we also perform experiments with phrase-level labels so as to have a fair and intuitive comparison with state-of-the-art baselines.  

We follow the same data split as described by Socher et al.~\cite{SST}: 8544/1101/2210 samples are used for training, validation and testing respectively in the 5-class task. The corresponding splits in the binary classification task are 6920/872/1821.
\vspace{-5mm}
\subsubsection{Results and Discussion}
\vspace{-2mm}
\paragraph{Evaluation of Classification Performance}
\begin{table}[!tb]
\caption{Classification accuracy (\%) on Stanford Sentiment TreeBank dataset when training with only the sentence-level labels. We conduct experiments on both binary and fine-grained (5-class) classification tasks. Note that our model outperforms all others in the task.}
\centering
\vspace{2mm}
\normalsize
\renewcommand\arraystretch{1.2}
\resizebox{0.90\linewidth}{!}{
\begin{tabular}{l|l|c|c}
\Xhline{1pt}
& Model & Binary & Fine-grained \\
\hline
\multirow{2}{*}{Graphical models} & HULM & 81.3 &44.1 \\
& HCRF & 84.8& 45.3\\
\hline
\multirow{1}{*}{\tabincell{l}{Syntactic compositions} }
& DAN-ROOT~\cite{SST_DeepUnordered} & 85.7 & 46.9\\
\hline
\multirow{3}{*}{Recurrent models}& Plain-RNN &83.9& 42.3\\
& GRU &85.4 & 46.7 \\
& LSTM &85.9 & 47.2\\
\hline
\multirow{1}{*}{Our model } & TAGM & \textbf{86.2} & \textbf{48.0} \\
\hline
\end{tabular}}
\label{table:SST1}
\end{table}
\begin{table}[!tb]
\caption{Classification accuracy ($\%$) on Stanford Sentiment TreeBank dataset  when  training with both phrase-level and sentence-level labels. Our TAGM achieves the best overall result.}
\centering
\vspace{2mm}
\normalsize
\renewcommand\arraystretch{1.2}
\resizebox{0.90\linewidth}{!}{
\begin{tabular}{l|l|c|c|c}
\Xhline{1.0pt}
& Model & Binary & Fine-grained  & \tabincell{l}{Overall \\ Performance}\\
\hline
\multirow{3}{*}{\tabincell{l}{Unordered \\ compositions} }& $\text{NBOW-RAND}$~\cite{SST_DeepUnordered} & 81.4 & 42.3 & 123.7\\
&$\text{NBOW}$~\cite{SST_DeepUnordered} & 83.6 & 43.6 & 127.2\\
& $\text{BiNB}$~\cite{SST_DeepUnordered}  & 83.1 & 41.9 & 125.0\\
\hline
\multirow{5}{*}{\tabincell{l}{Syntactic \\ compositions} }& $\text{RecNN}$~\cite{Socher_2011b}  & 82.4 & 43.2 & 125.6\\
& $\text{RecNTN}$~\cite{SST} & 85.4 & 45.7 & 131.1\\
& $\text{DRecNN}$~\cite{DRecNN} & 86.6 & 49.8 & 136.4\\
& $\text{DAN}$~\cite{SST_DeepUnordered} & 86.3 & 47.7 & 134.0\\
& $\text{TreeLSTM}$~\cite{TREE-LSTM} & 86.9 & $\mathbf{50.6}$ & 137.5\\
& $\text{CNN-MC}$~\cite{CNN-MC}  & $\mathbf{88.1}$& 47.4 & 135.5 \\
& $\text{PVEC}$~\cite{PVEC} & $87.8$ & 48.7 & 136.5\\
\hline
\multirow{1}{*}{Our model }  & $\text{TAGM}$ & 87.6 & 50.1 & \textbf{137.7}\\
\hline
\end{tabular}}
\label{table:SST2}
\end{table}
We conduct two sets of experiments to evaluate the performance of our model in comparison with the baseline models. Since our model is designed for unsegmented and possibly noisy sequences modeling, it is more suitable to only use sentence-level labels, although phrase-level labels are also provided in SST dataset. Table~\ref{table:SST1} shows the experimental results of several sequential models trained with only sentence-level labels.  Our model achieves the best result in both binary classification task and fine-grained (5-class) task. LSTM and GRU outperform plain-RNN model due to the information-filtering capability performed by additional gates. It is worth mentioning that our model achieves  better performance than LSTM with only half the hidden parameters.
\begin{figure*}[!t]
\begin{minipage}[l]{0.94\textwidth} 
\centering
	\includegraphics[width=0.99\linewidth]{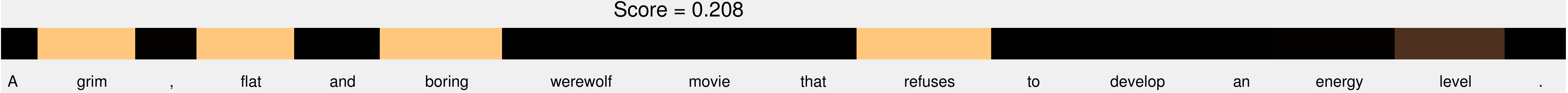} \\
	\includegraphics[width=0.99\linewidth]{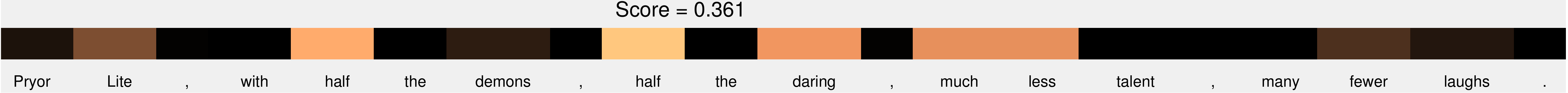} \\ 
	\includegraphics[width=0.99\linewidth]{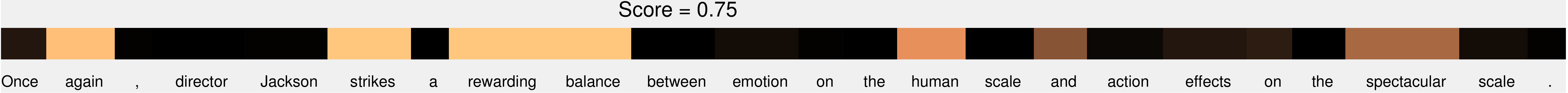} \\
    \includegraphics[width=0.99\linewidth]{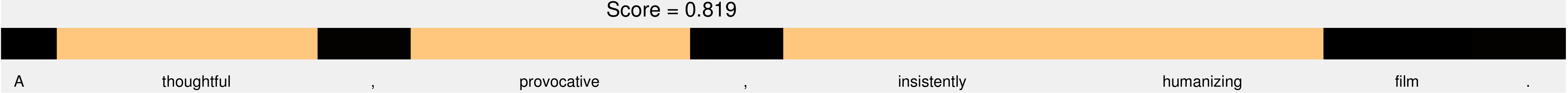}  
\end{minipage}
\begin{minipage}[l]{0.05\textwidth} 
	 \includegraphics[width=0.85\linewidth, left]{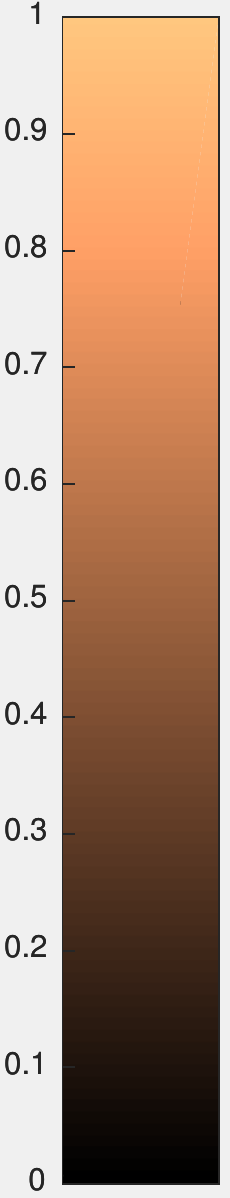}
\end{minipage}
\begin{minipage}[c]{\textwidth} 
\includegraphics[width=\linewidth]{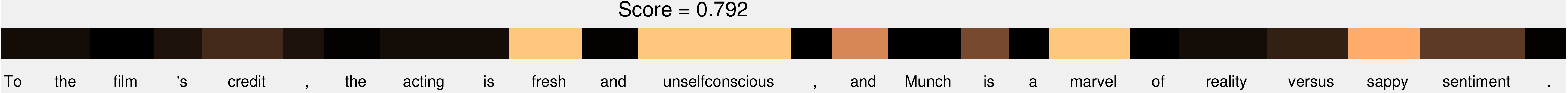}\\   
\includegraphics[width=\linewidth]{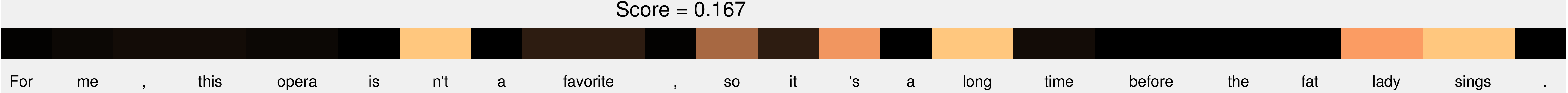}\\             
\includegraphics[width=\linewidth]{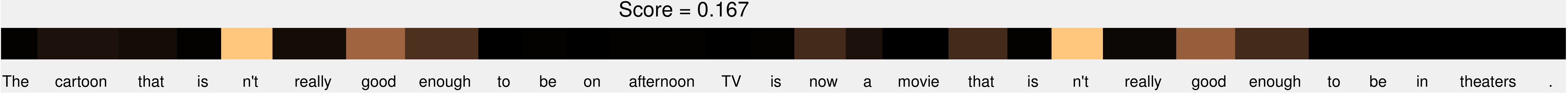}\\      
\centering	(a) \quad Correct predictions. \vspace{1mm}\\    
   \includegraphics[width=\linewidth]{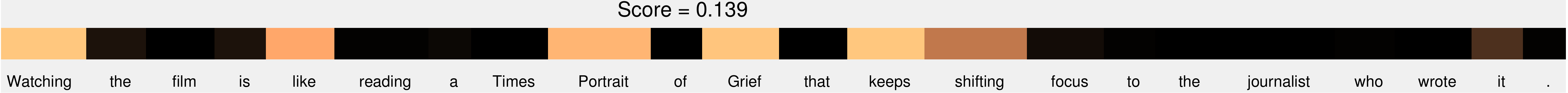} \\    
  \includegraphics[width=\linewidth]{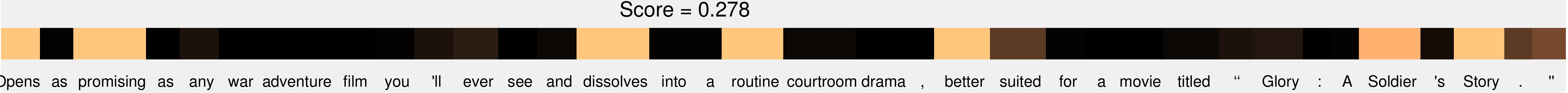} \\  
\centering   (b) \quad Wrong predictions.   
\end{minipage}
\vspace{1mm}
   \caption{ The visualization of attention weights of Recurrent Attention Model: (a) correct predictions and (b) wrong predictions. The scores displayed are the groundtruth label indicating the writer's overall sentiment for this review. Darker color indicates smaller scores.}
\label{fig:SST}
\end{figure*}

To have a fair comparison with the existing sentiment analysis models, we conduct the second set of experiments with both sentence-level and phrase-level labels. The results are presented in Table~\ref{table:SST2}. It shows that our model outperforms most of the existing models and achieves comparable accuracy with the state-of-the-art results. Our TAGM model actually obtains overall best results considering both binary and fine-grained cases. This is an encouraging result, in particular, since our model is not specifically designed towards NLP tasks.
\vspace{-3mm}
\paragraph{Sequence Salience Detection}
In order to investigate the performance of salience detection by our TAGM model on Sentiment dataset (SST), we visualize the calculated attention weights for each word in the test sentences. Group (a) in Figure~\ref{fig:SST} presents a number of examples that are predicted correctly by our model in the binary-classification task.  
It shows that our model is able to successfully  capture the key sentimental words and omit irrelevant words, even for the sentences with complicated syntax. We also test the examples that include negated expressions. As shown in the last two sentences of group (a),  our model can deal with them very well. 
We also investigate the samples our model fails to predict the correct sentiment label (see Figure~\ref{fig:SST}b).
\vspace{-1mm}
\subsection{Event recognition Experiments}
\vspace{-2mm}
We subsequently conduct experiments for video event recognition to evaluate our model on the visual modality. 
\vspace{-2mm}
\subsubsection{Dataset}
\vspace{-2mm}
Columbia Consumer Video (CCV) Database~\cite{CCV} is an unconstrained video database collected from YouTube videos without any post-editing. It consists of 9317 web videos with average duration of 80 seconds (210 hours in total). Except for some negative background videos, each video is manually annotated into one or more of 20 semantic categories such as `basketball', `ice skating', `biking', `birthday' and so on. It is a very challenging database due to the many noisy and irrelevant segments contained inside these videos. 
\begin{figure*}[!tb]
\begin{center}
\begin{tabular}{cccccccccc}
   \includegraphics[width=0.093\linewidth]{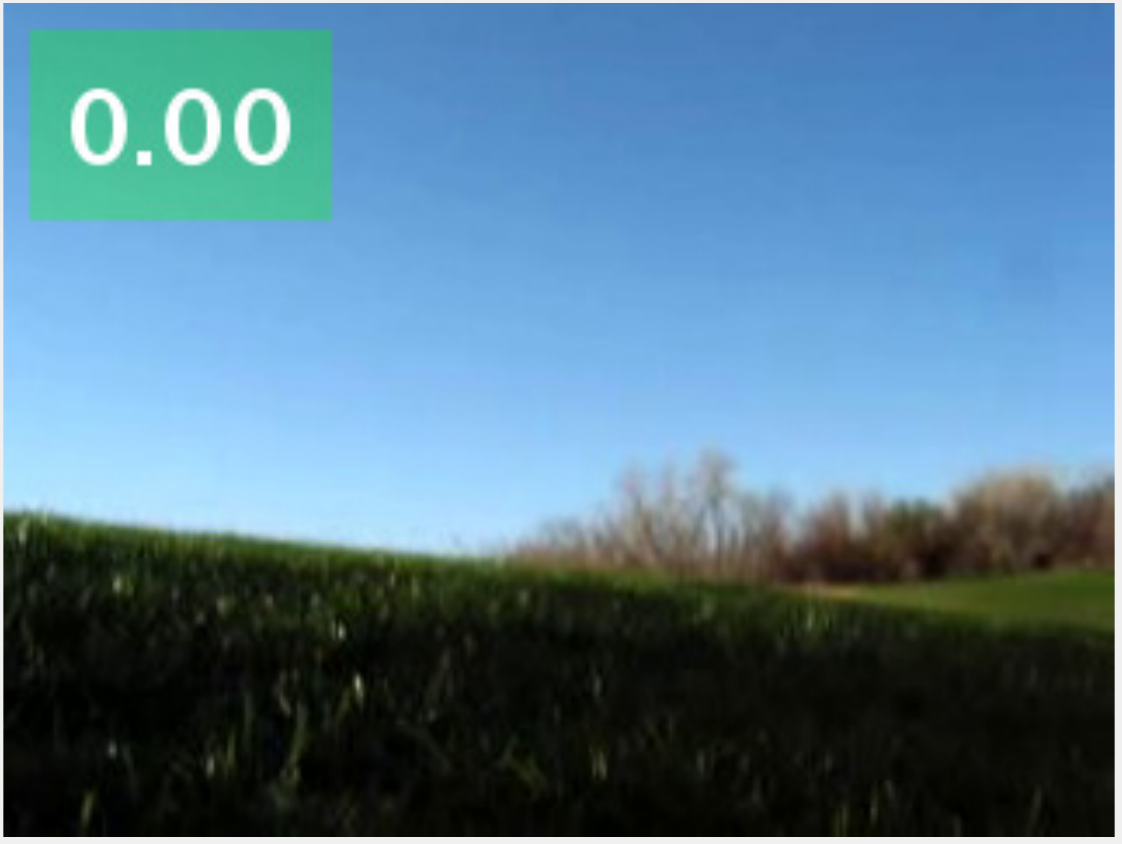} 
   \includegraphics[width=0.093\linewidth]{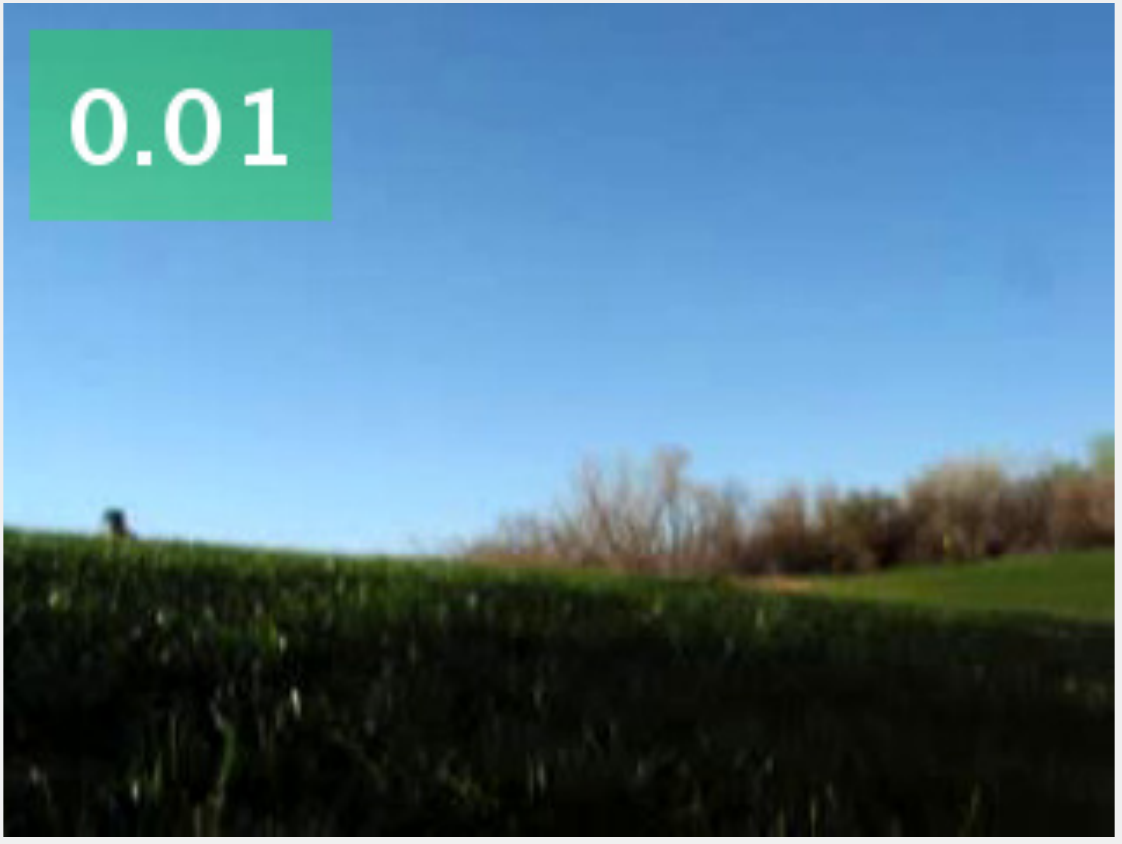}   
   \includegraphics[width=0.093\linewidth]{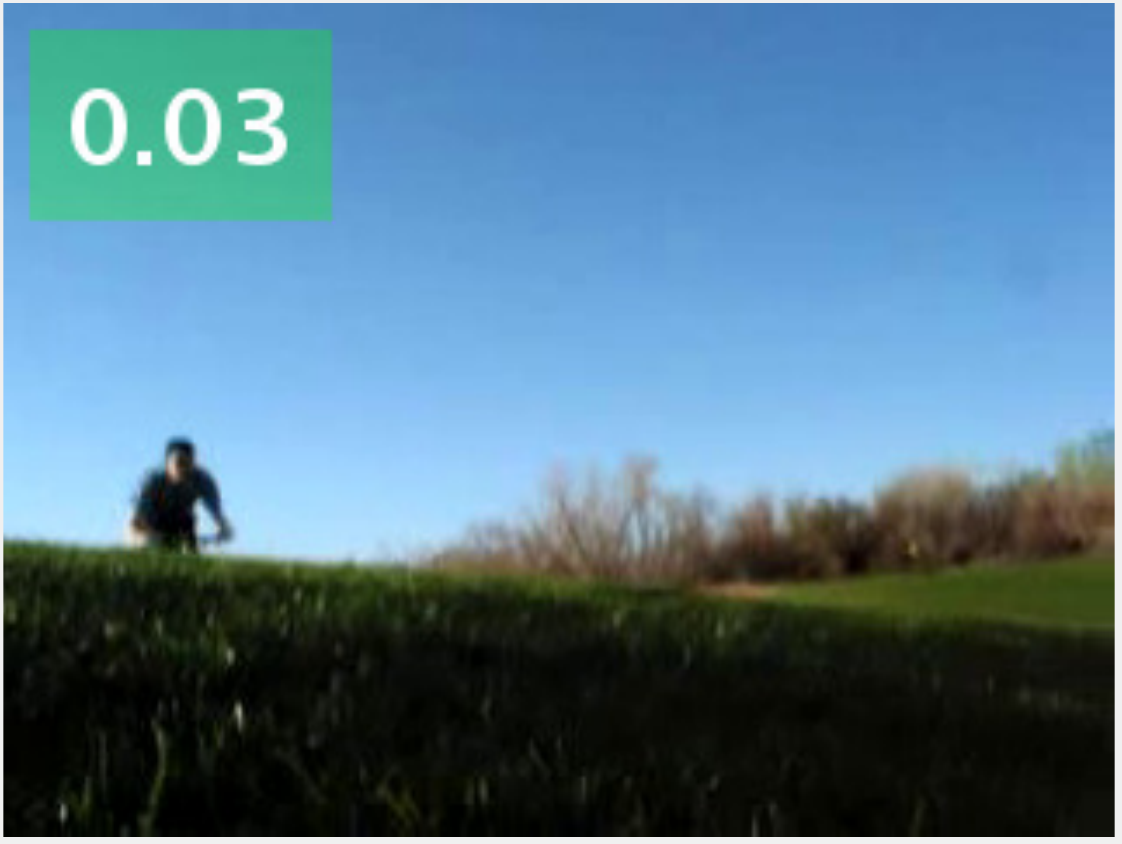}   
   \includegraphics[width=0.093\linewidth]{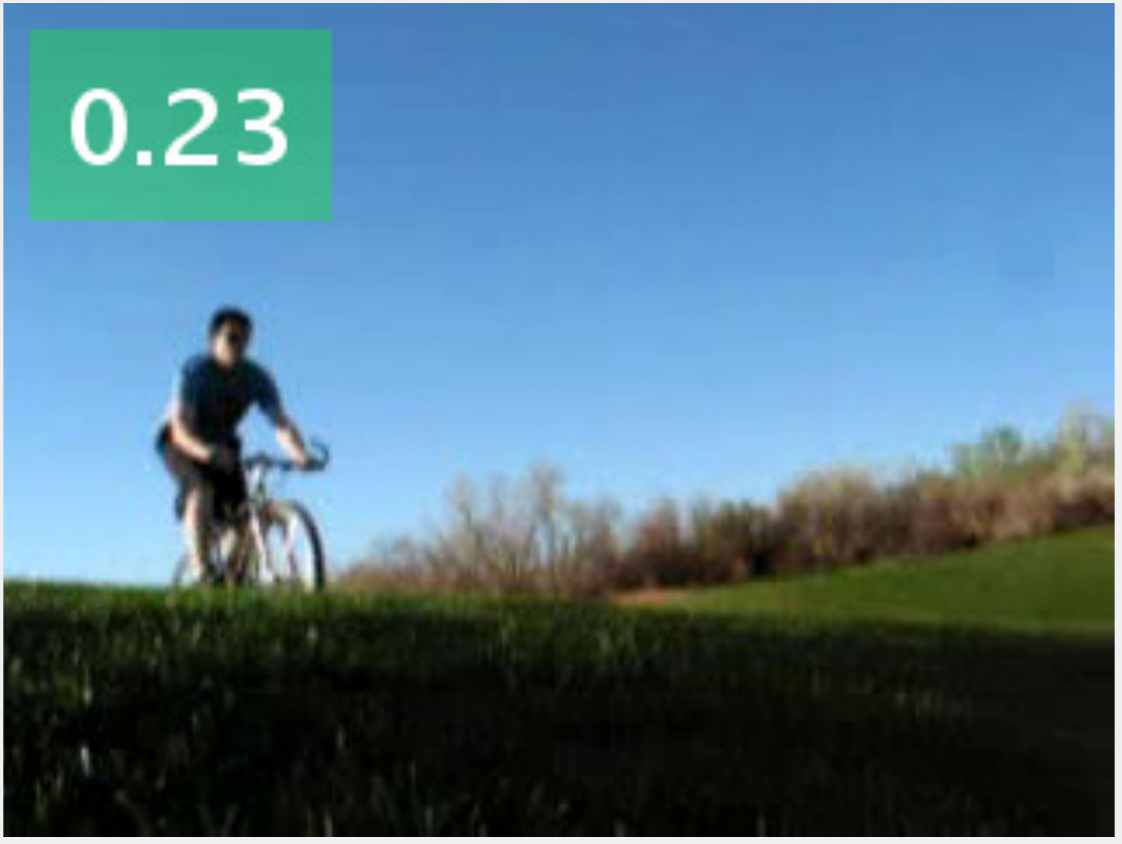}   
   \includegraphics[width=0.093\linewidth]{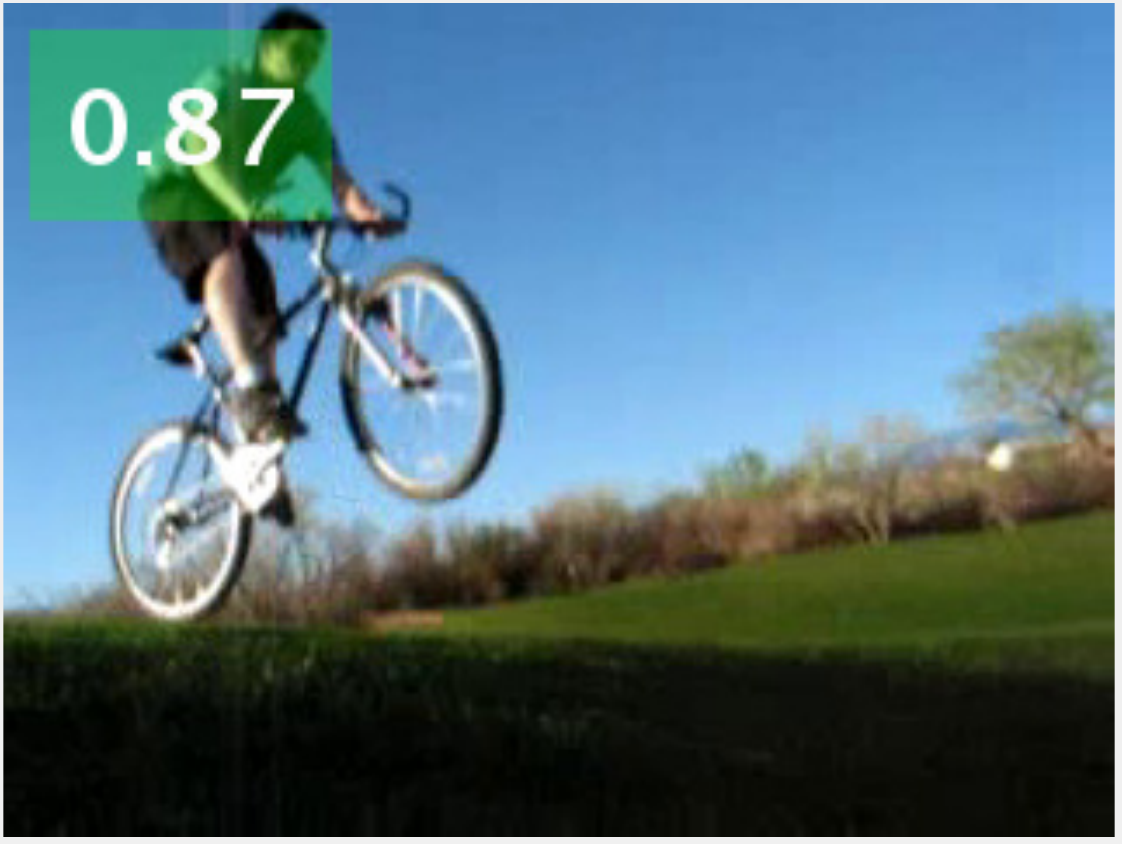}    
   \includegraphics[width=0.093\linewidth]{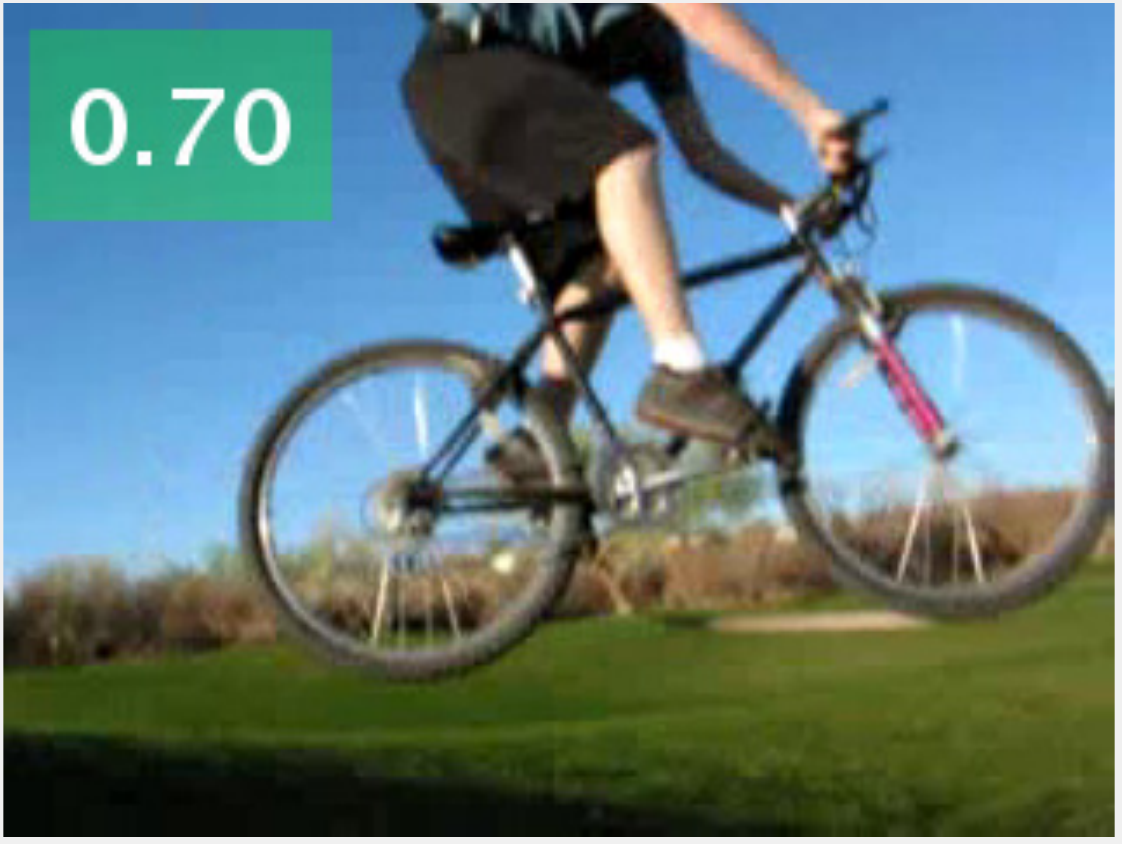}  
   \includegraphics[width=0.093\linewidth]{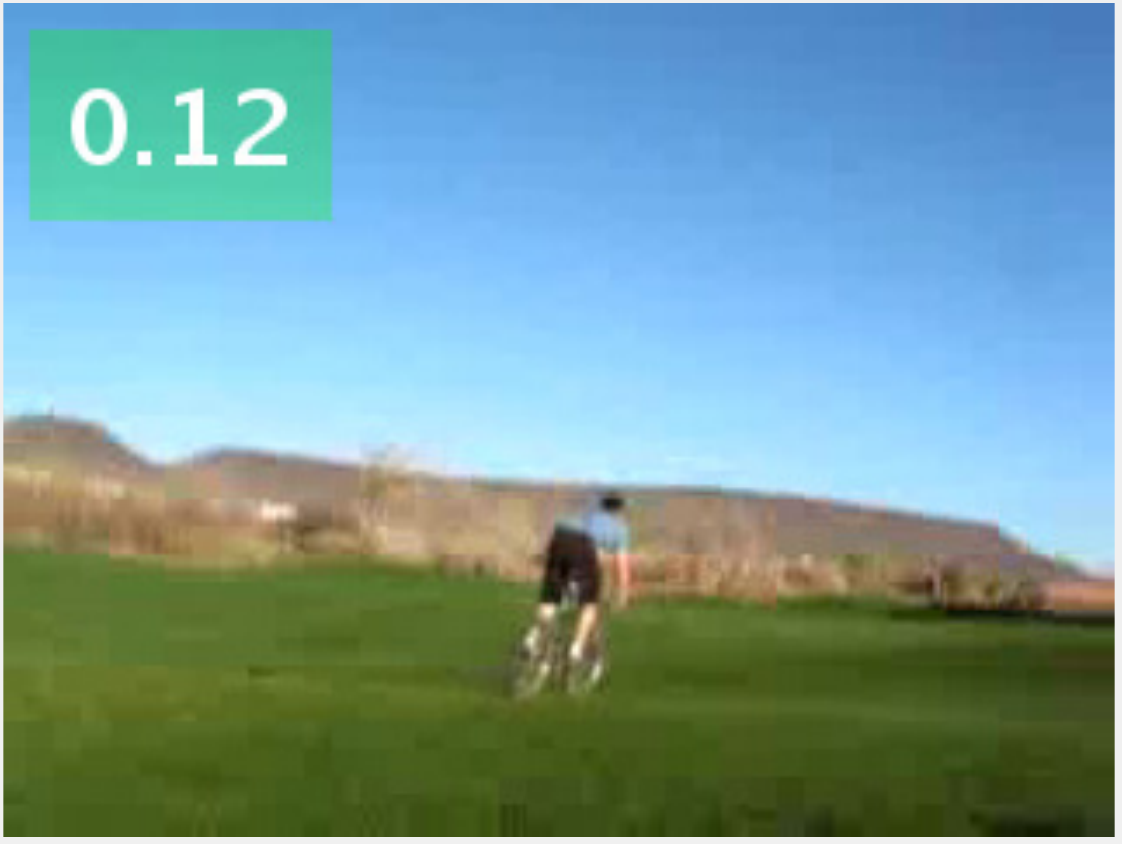}    
   \includegraphics[width=0.093\linewidth]{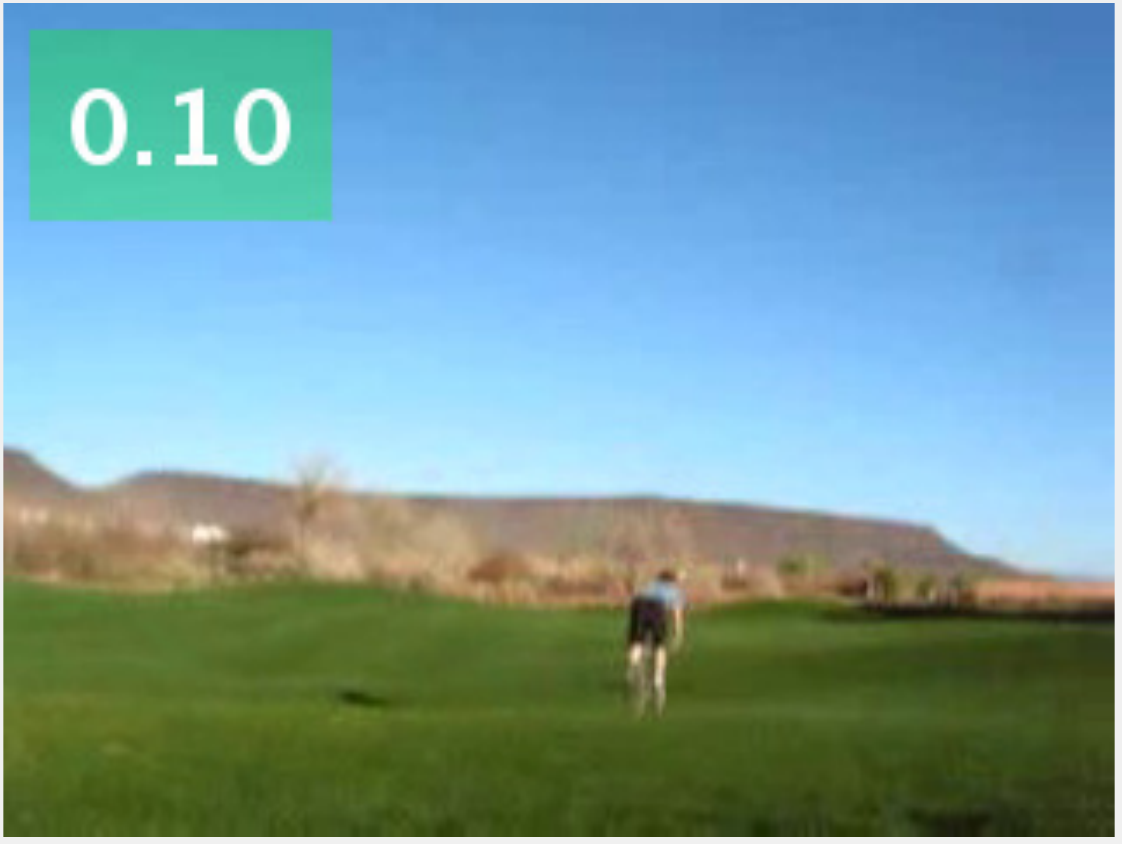}      
   \includegraphics[width=0.093\linewidth]{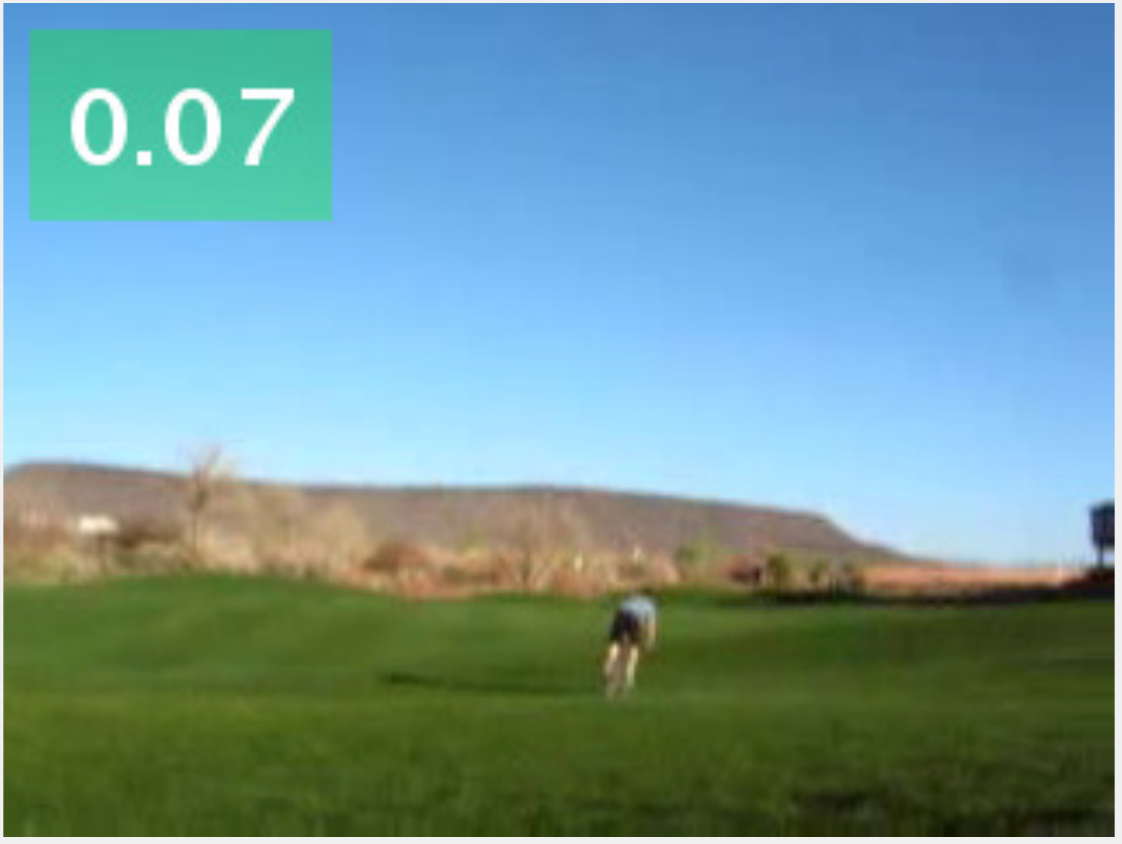}      
   \includegraphics[width=0.093\linewidth]{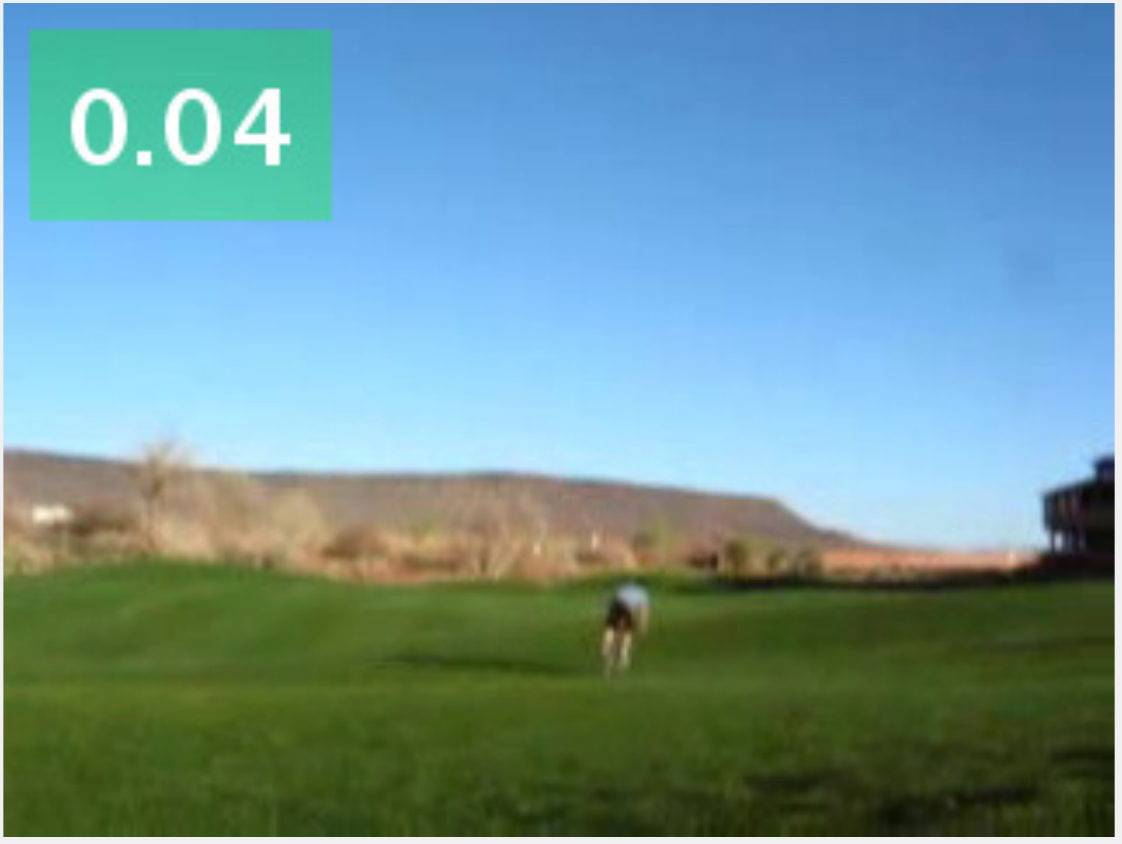} \\
   \small{Event: biking}    \vspace{0.2cm}\\
   \includegraphics[width=0.093\linewidth]{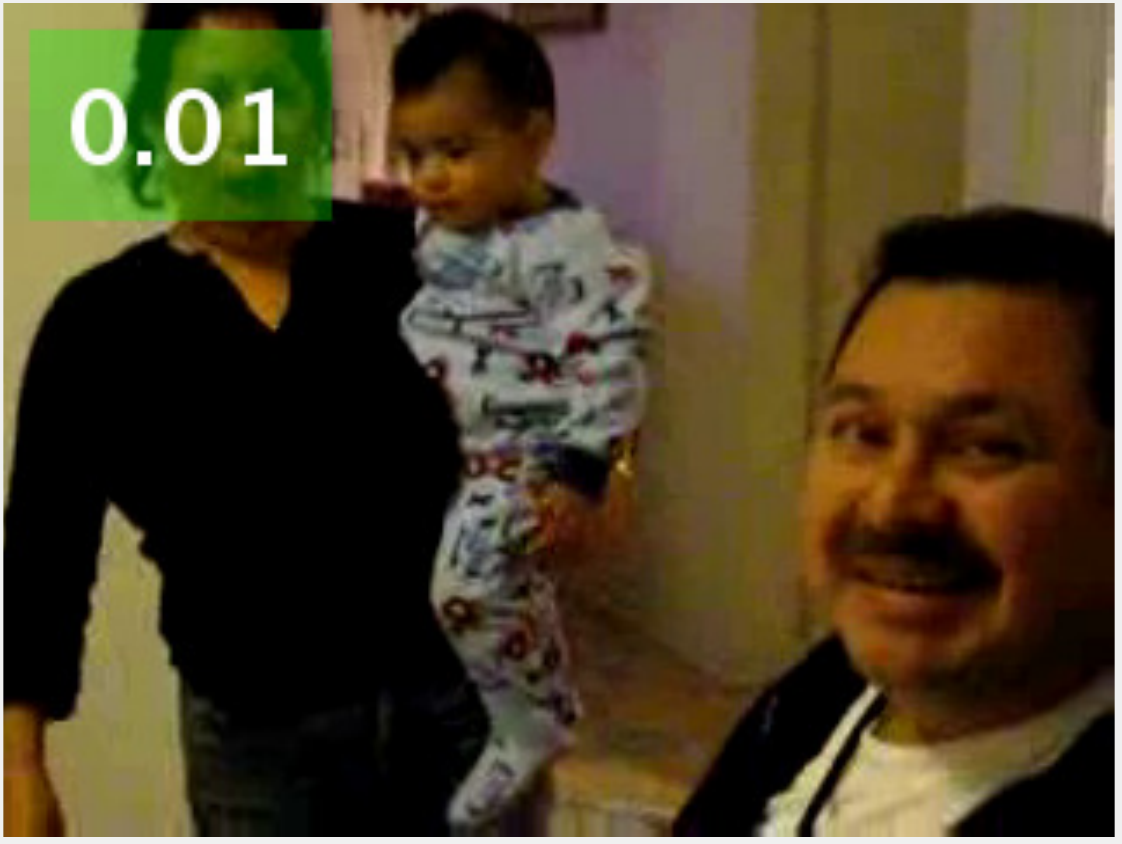}
   \includegraphics[width=0.093\linewidth]{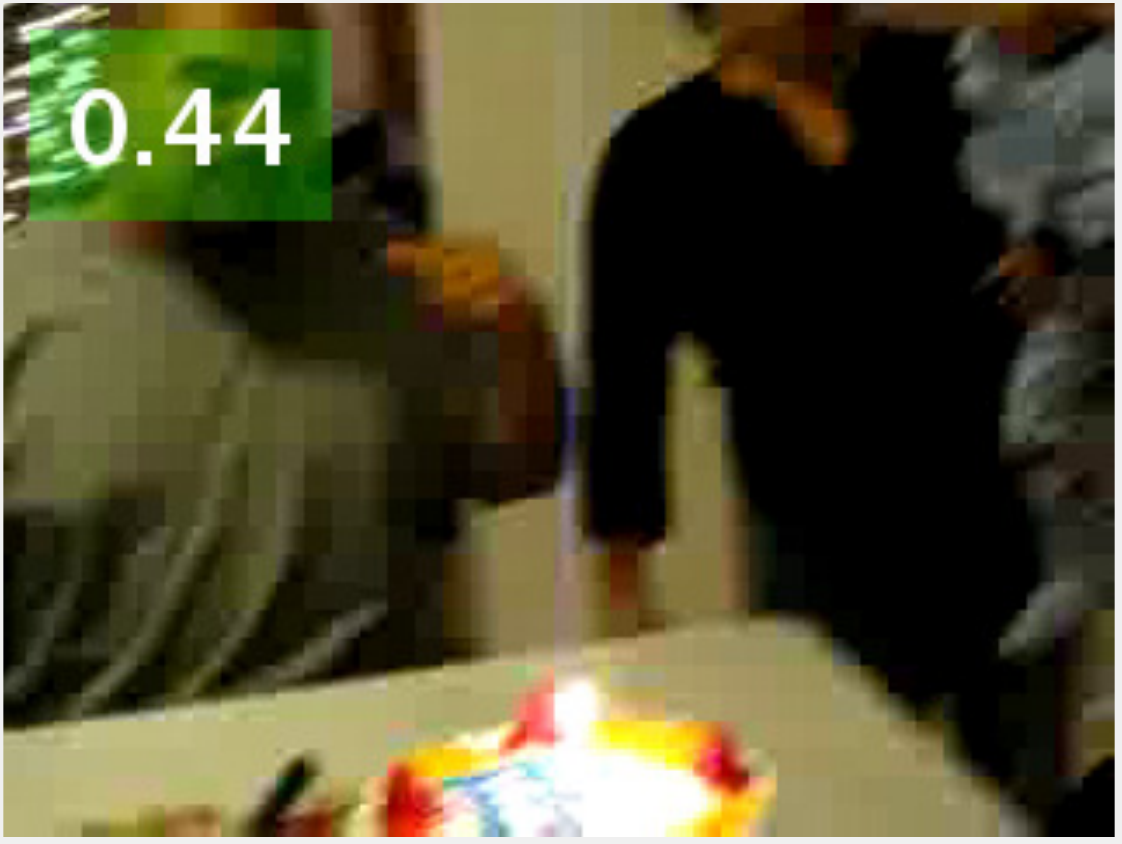}
   \includegraphics[width=0.093\linewidth]{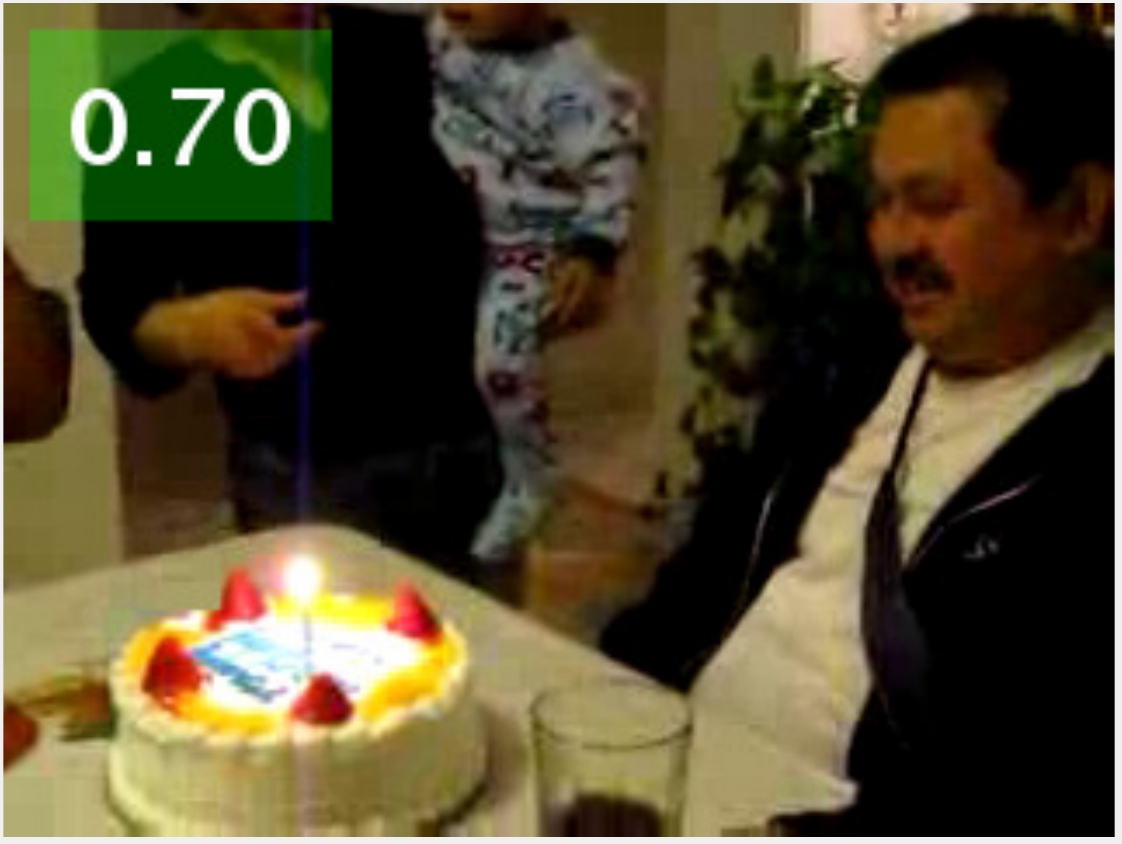}   
   \includegraphics[width=0.093\linewidth]{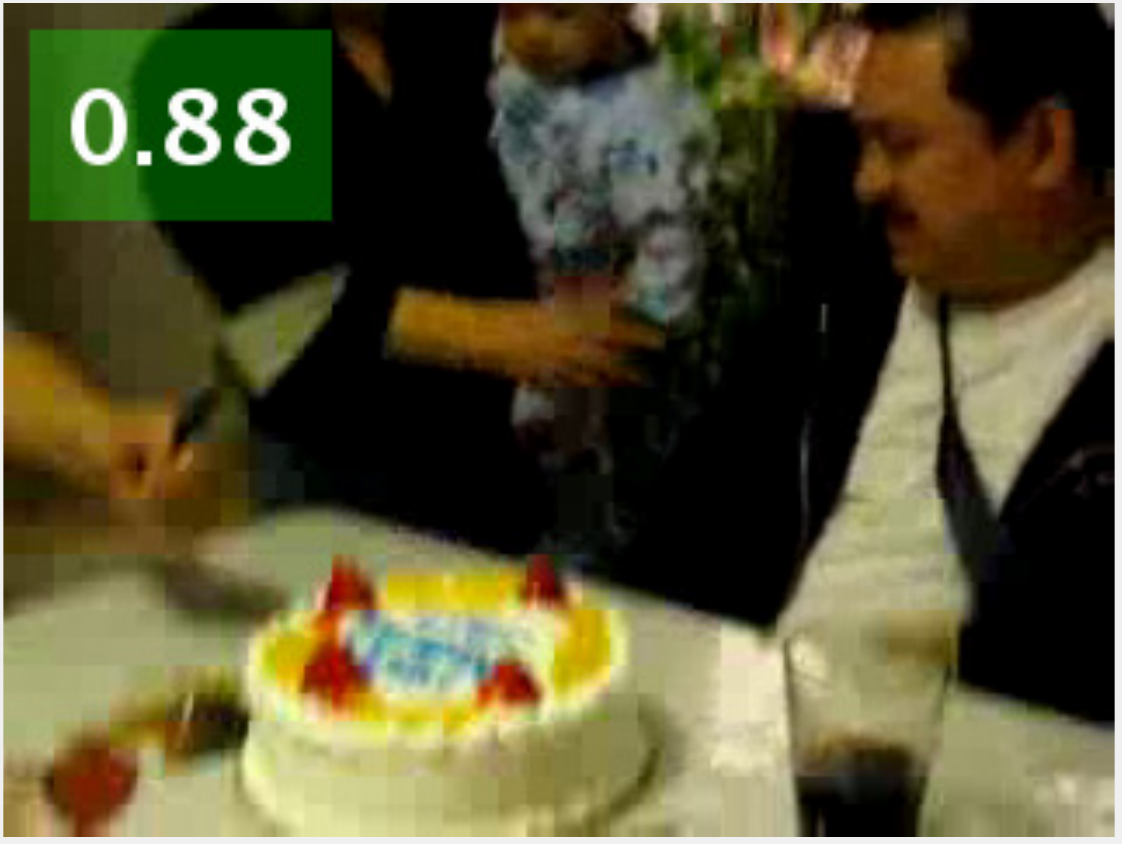}   
   \includegraphics[width=0.093\linewidth]{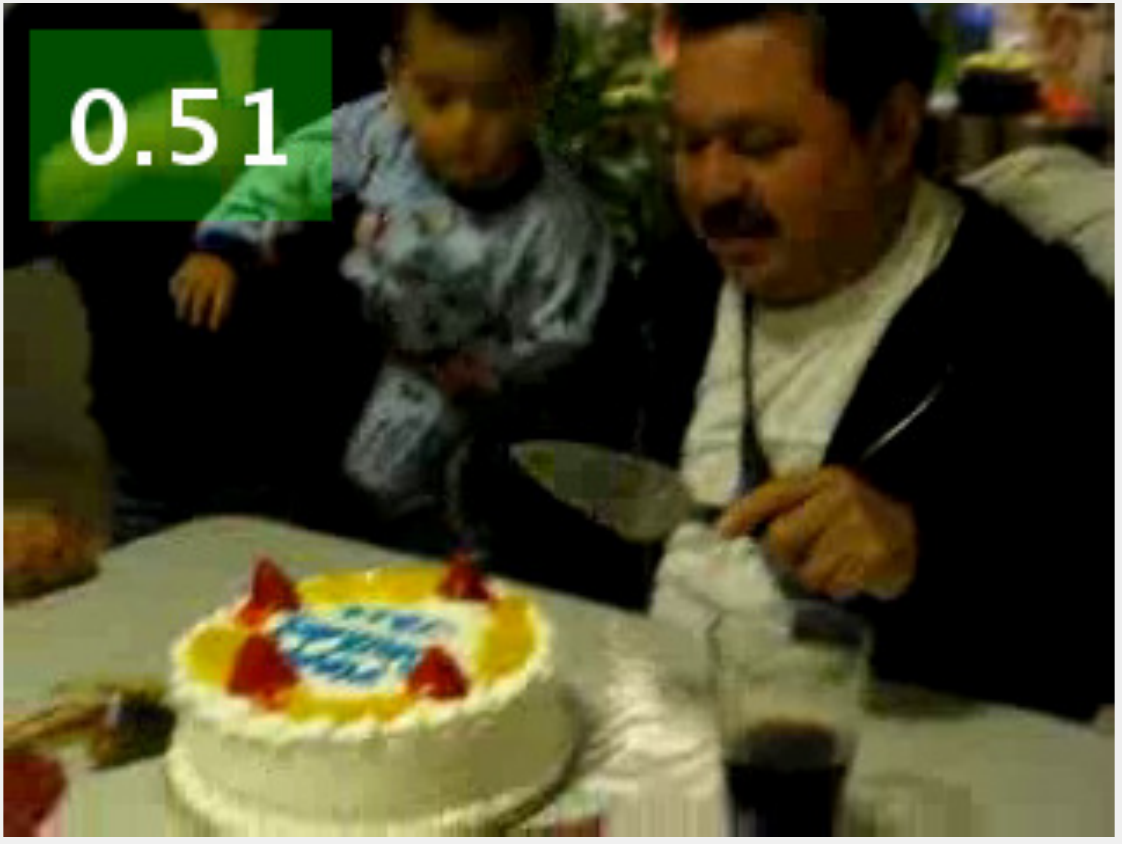}  
   \includegraphics[width=0.093\linewidth]{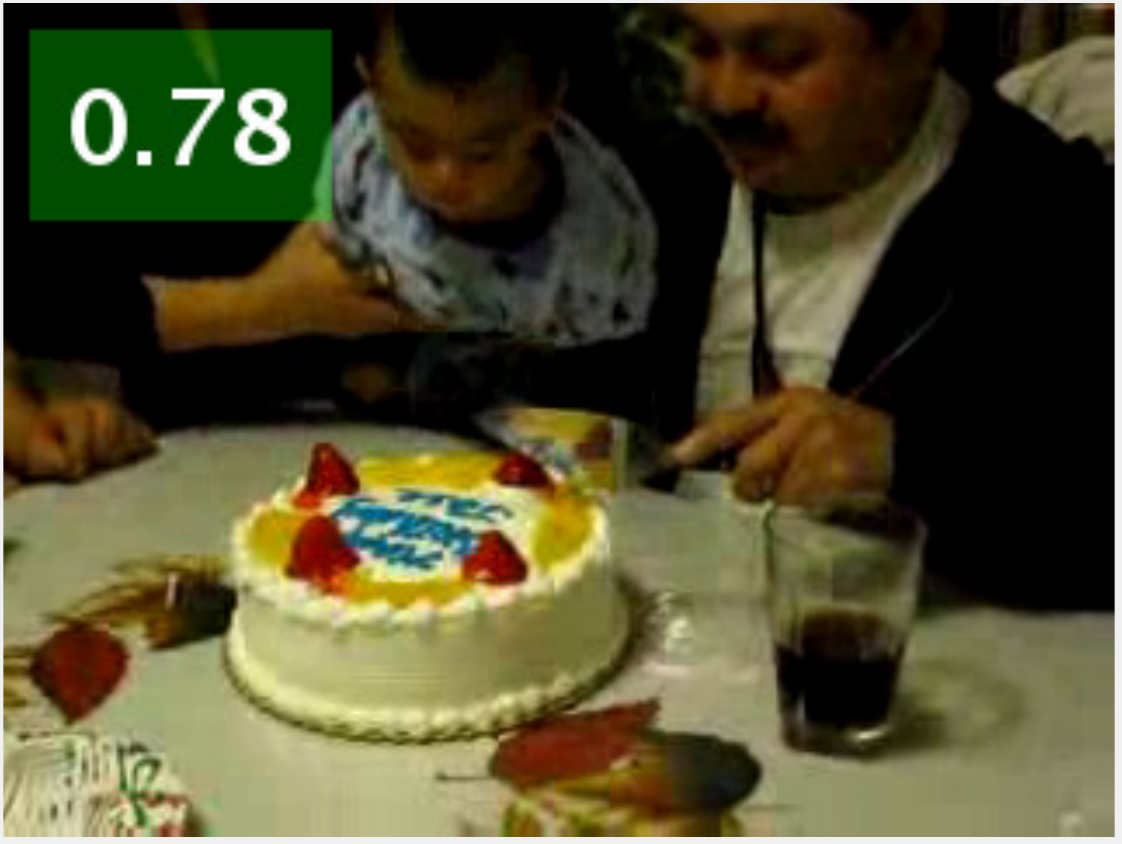}    
   \includegraphics[width=0.093\linewidth]{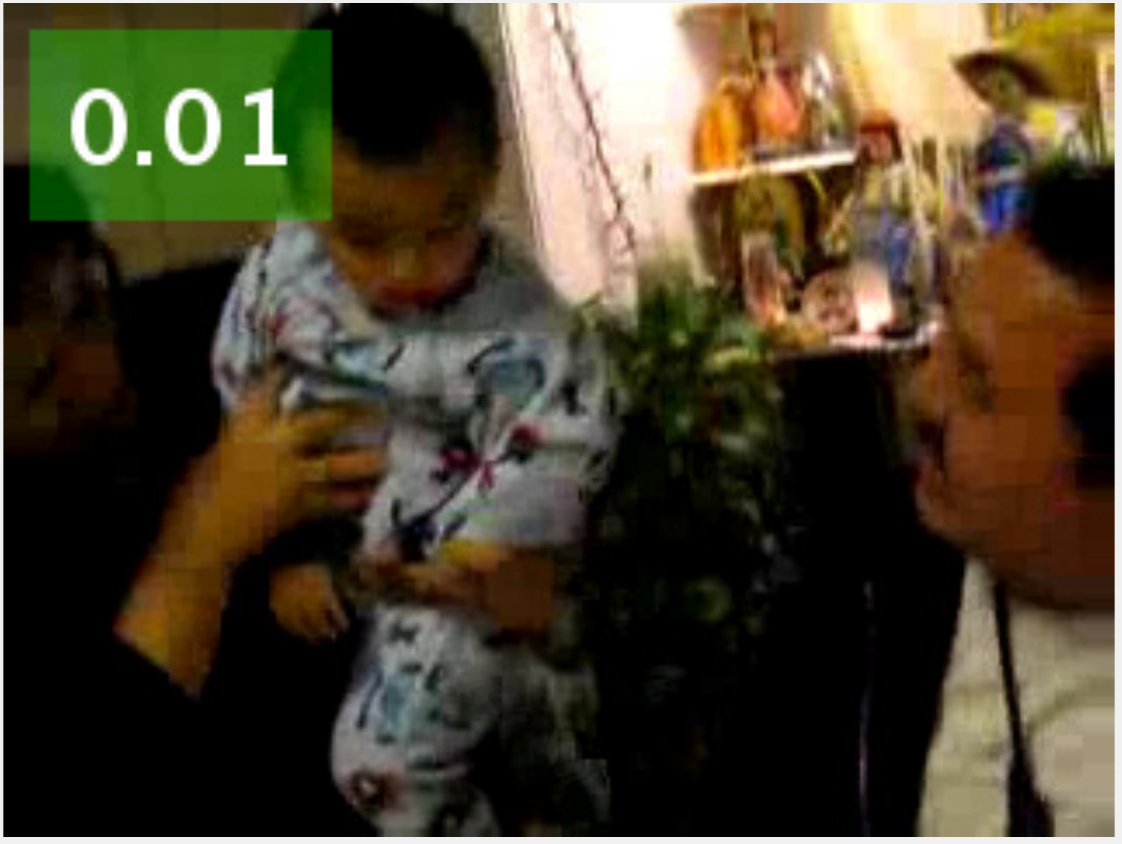}      
   \includegraphics[width=0.093\linewidth]{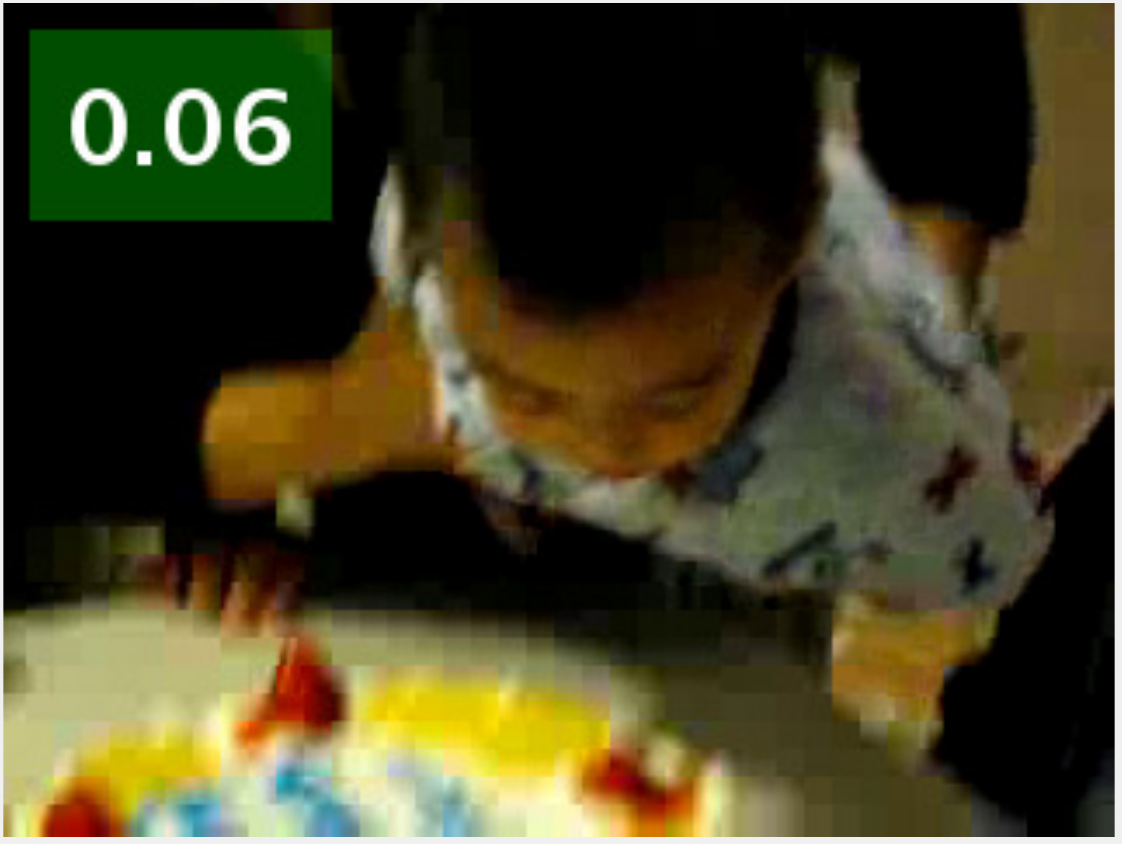}      
   \includegraphics[width=0.093\linewidth]{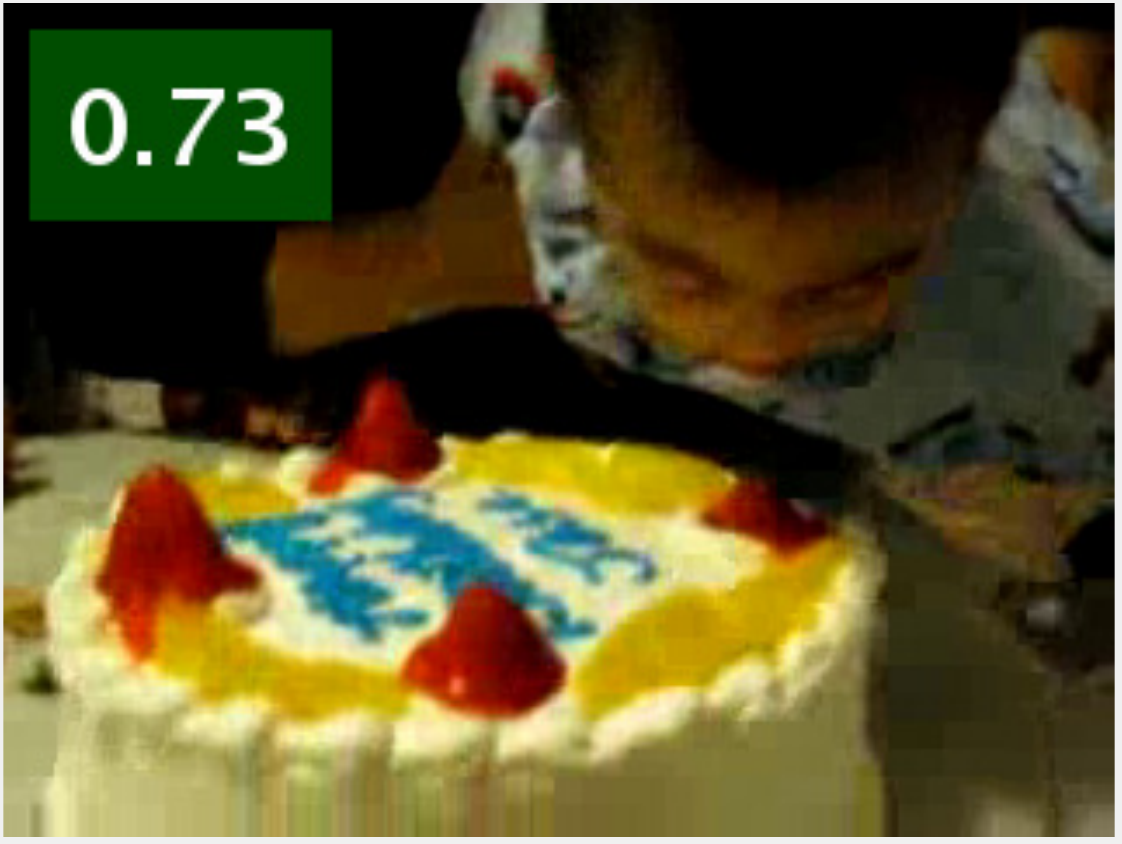}         
   \includegraphics[width=0.093\linewidth]{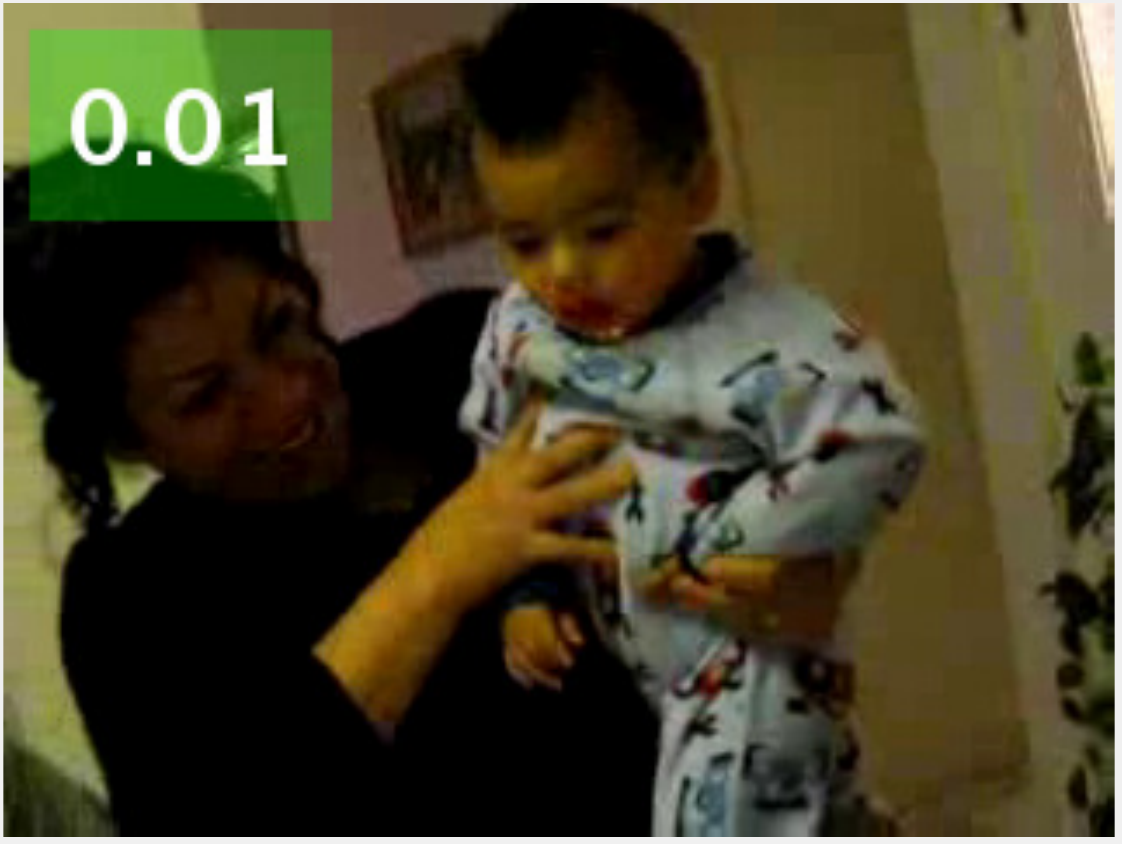} \\
   \small{Event: birthday} \vspace{0.2cm}\\
   \includegraphics[width=0.093\linewidth]{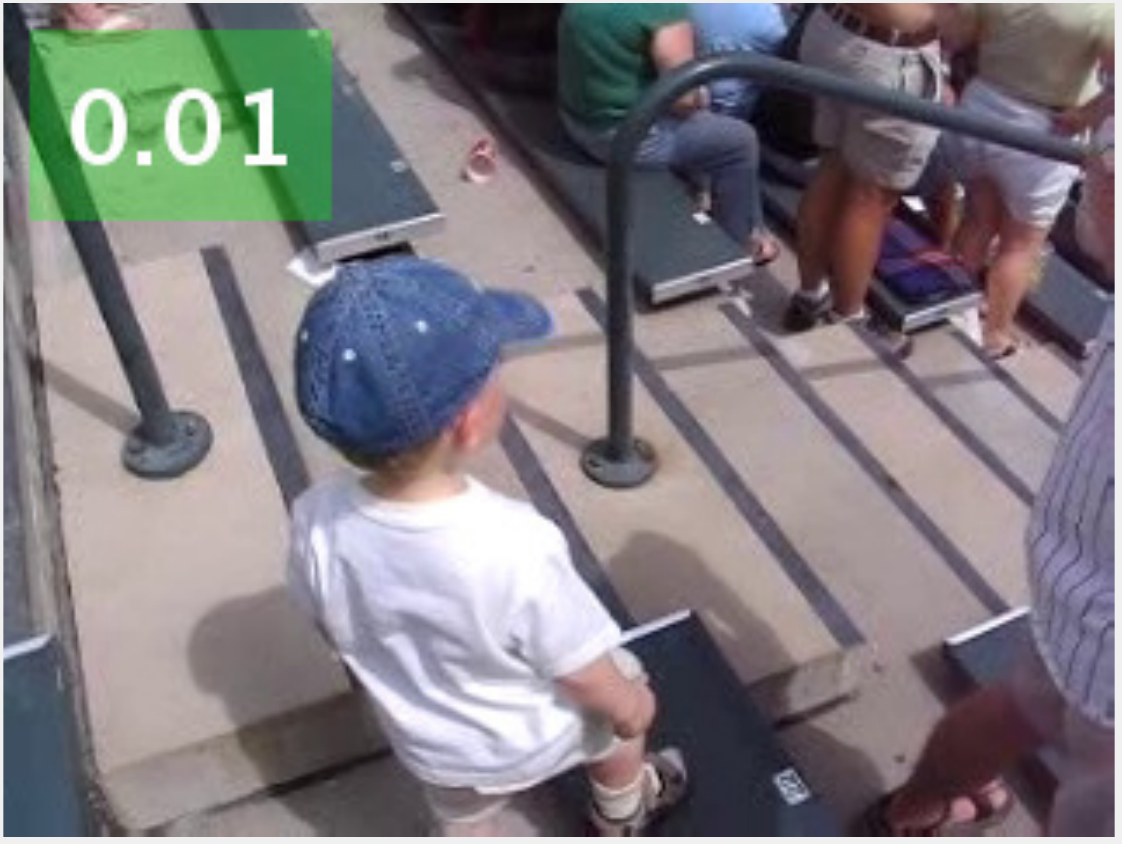}
   \includegraphics[width=0.093\linewidth]{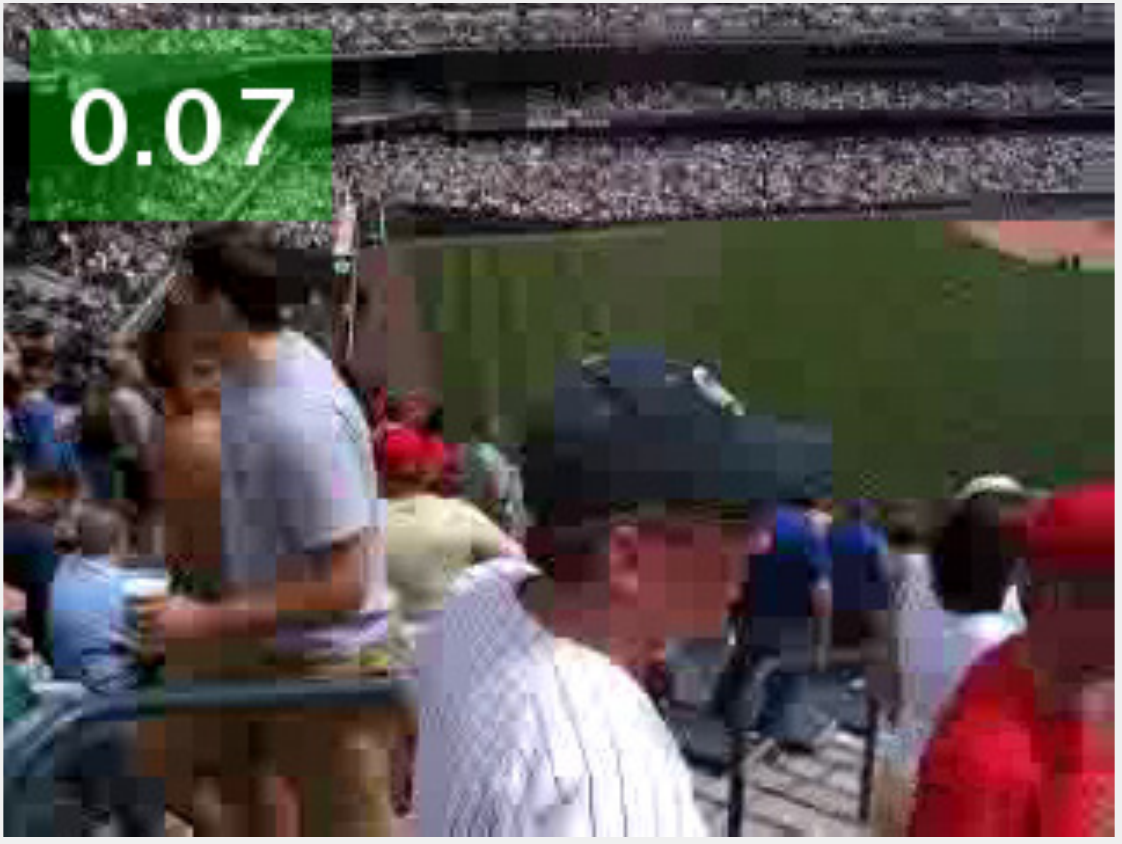}
   \includegraphics[width=0.093\linewidth]{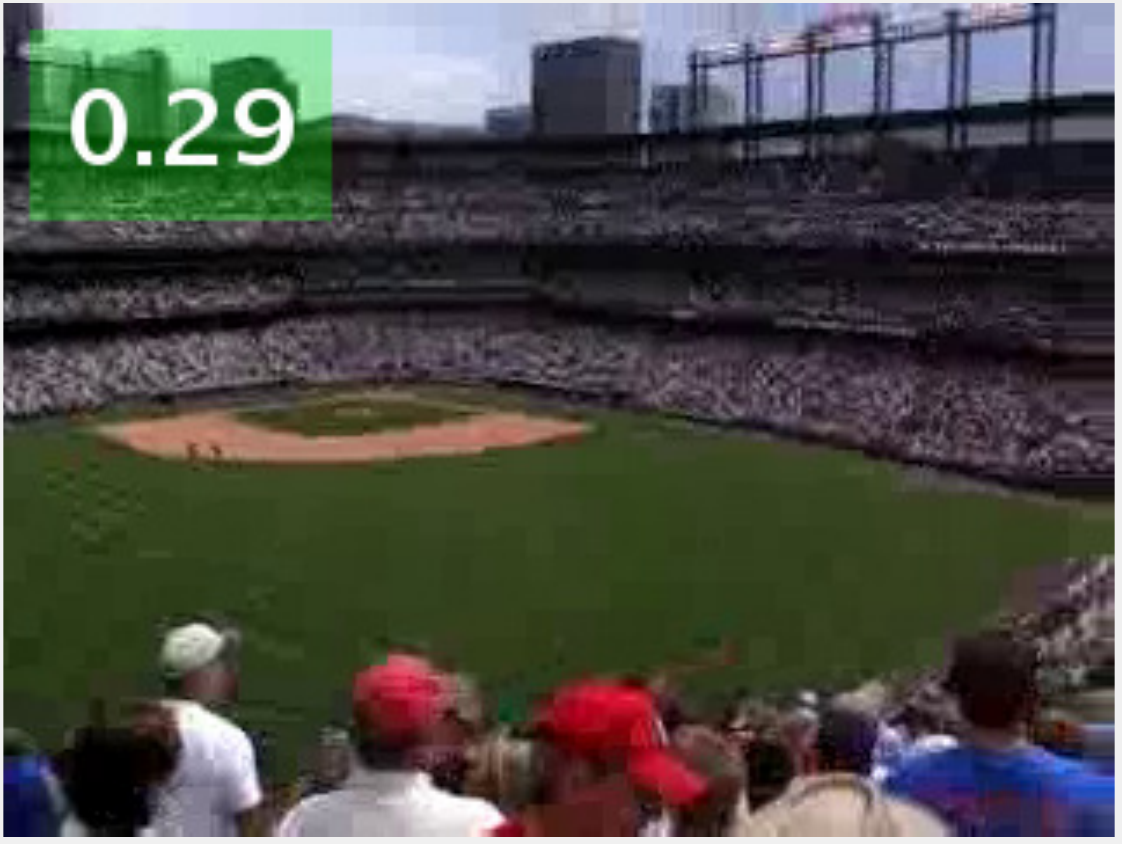}  
   \includegraphics[width=0.093\linewidth]{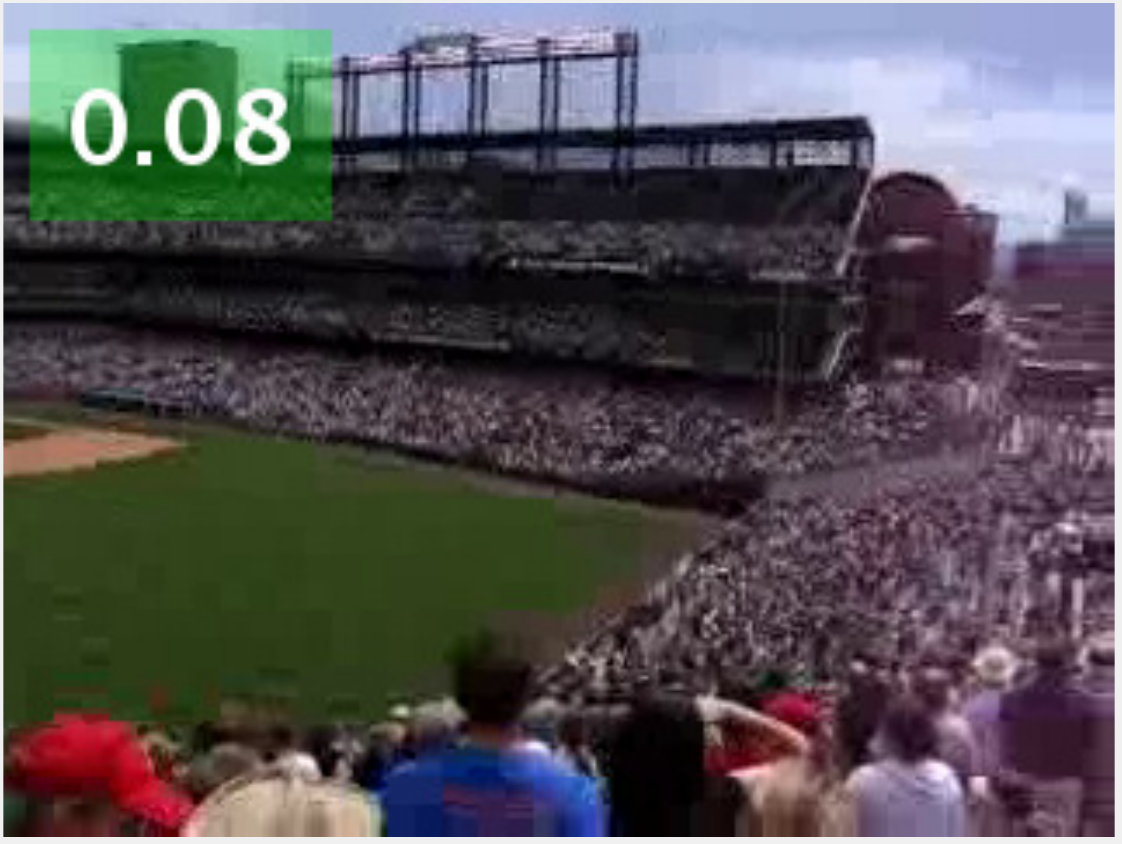}  
   \includegraphics[width=0.093\linewidth]{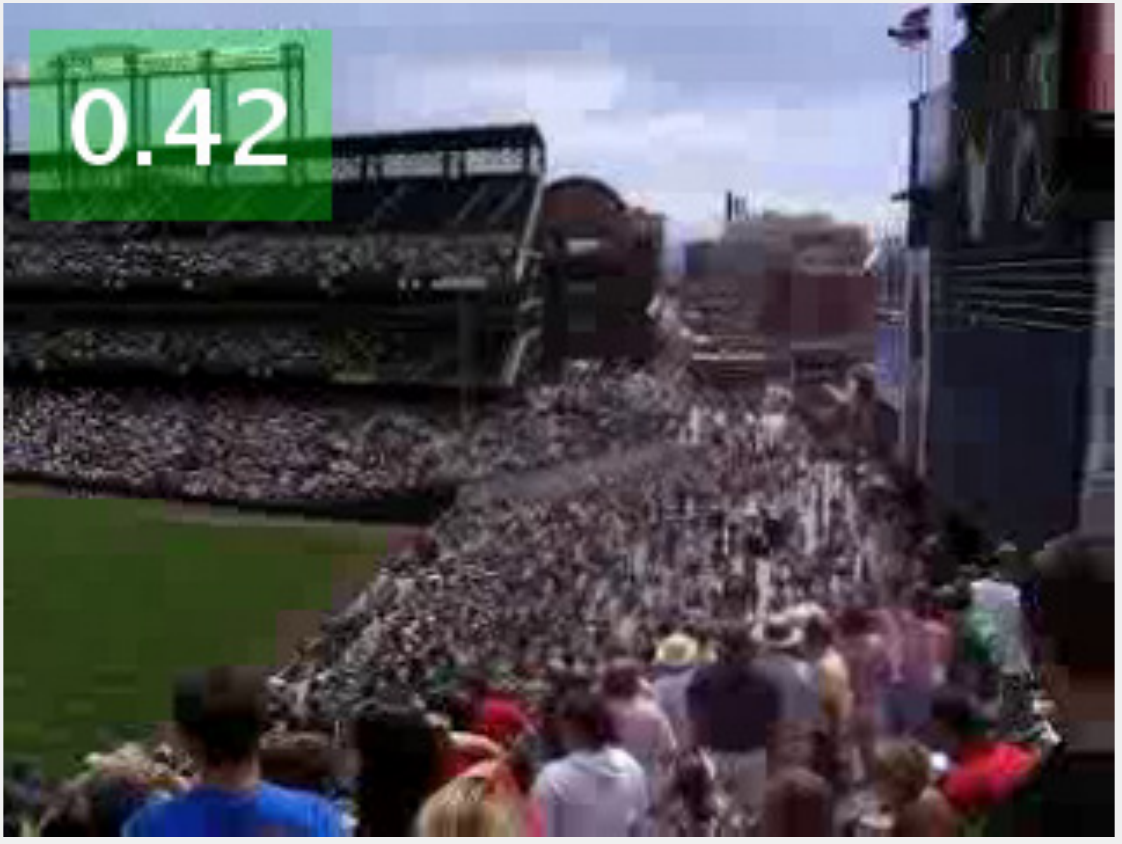} 
   \includegraphics[width=0.093\linewidth]{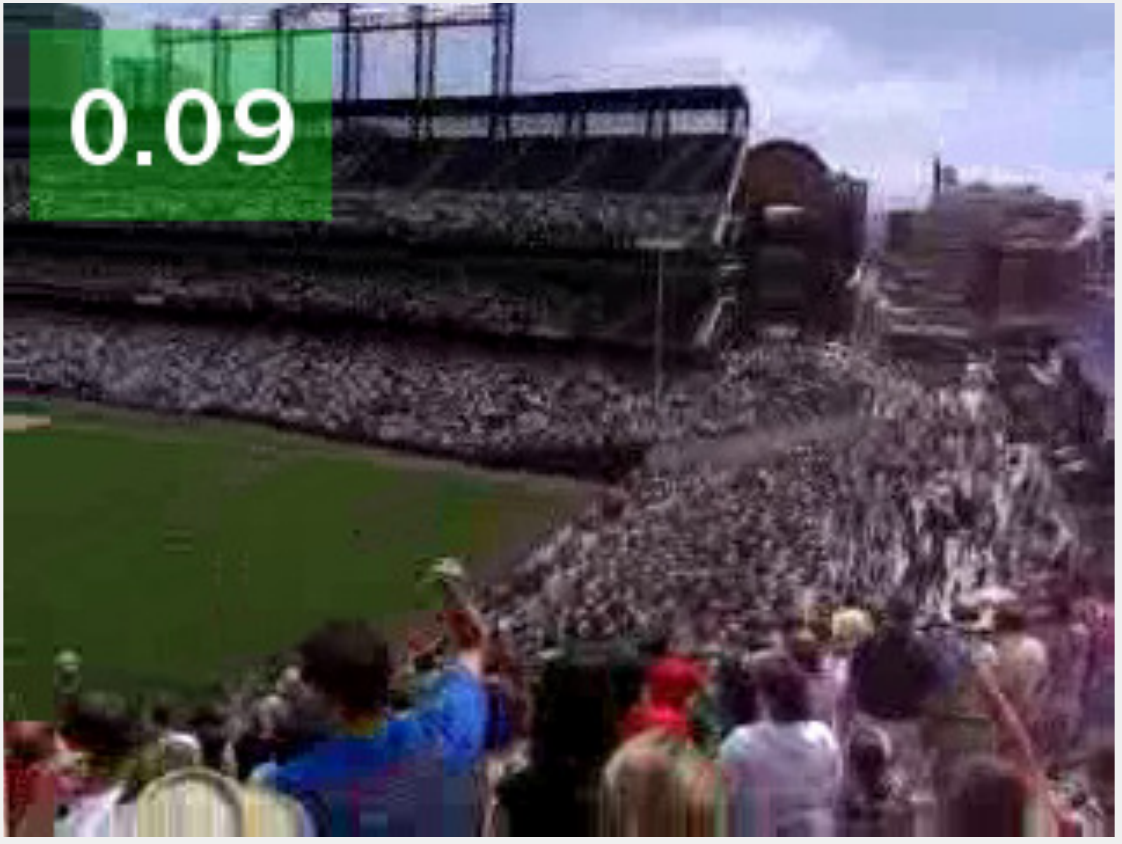}   
   \includegraphics[width=0.093\linewidth]{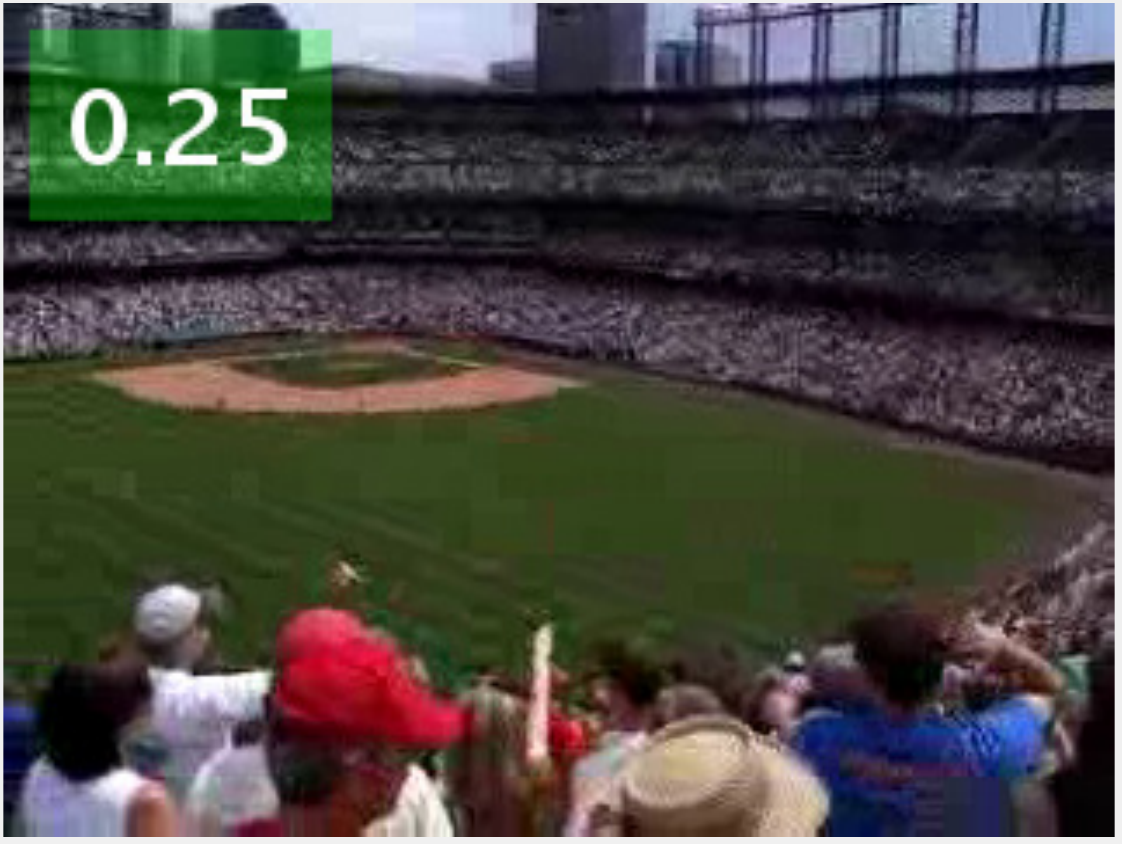}  
   \includegraphics[width=0.093\linewidth]{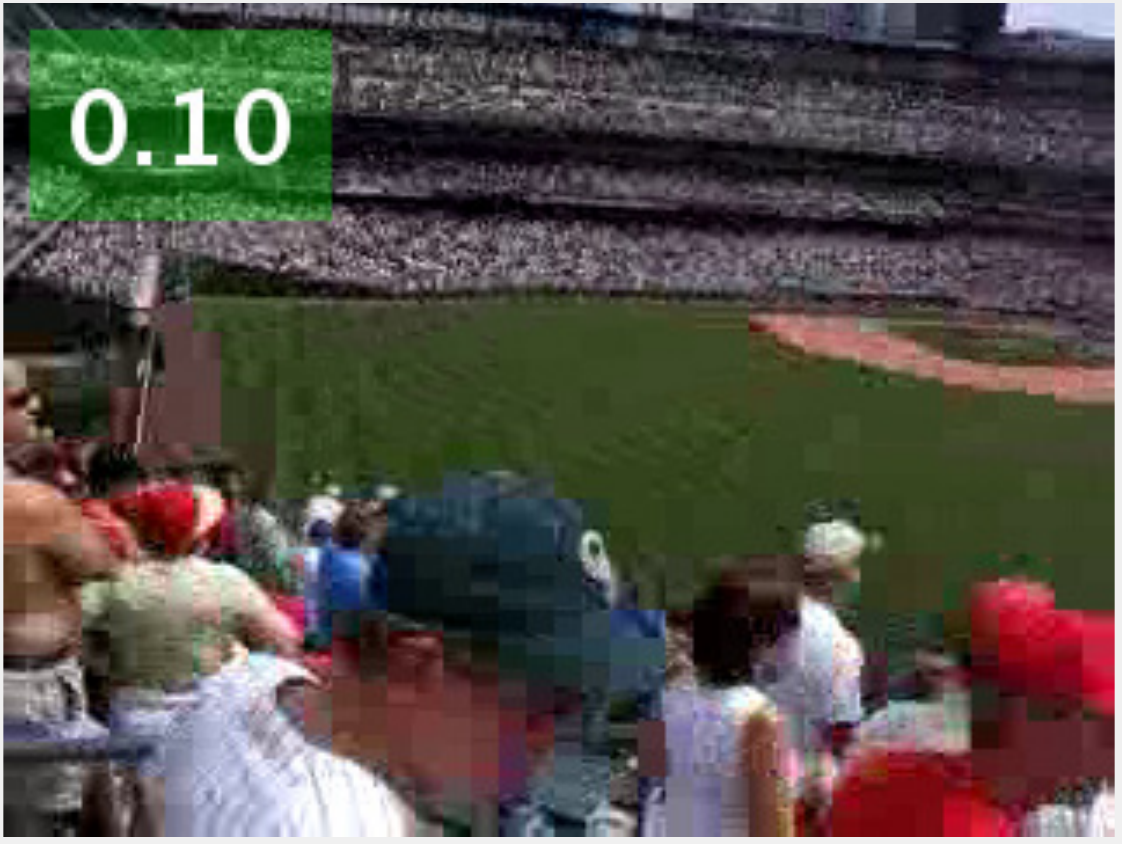}  
   \includegraphics[width=0.093\linewidth]{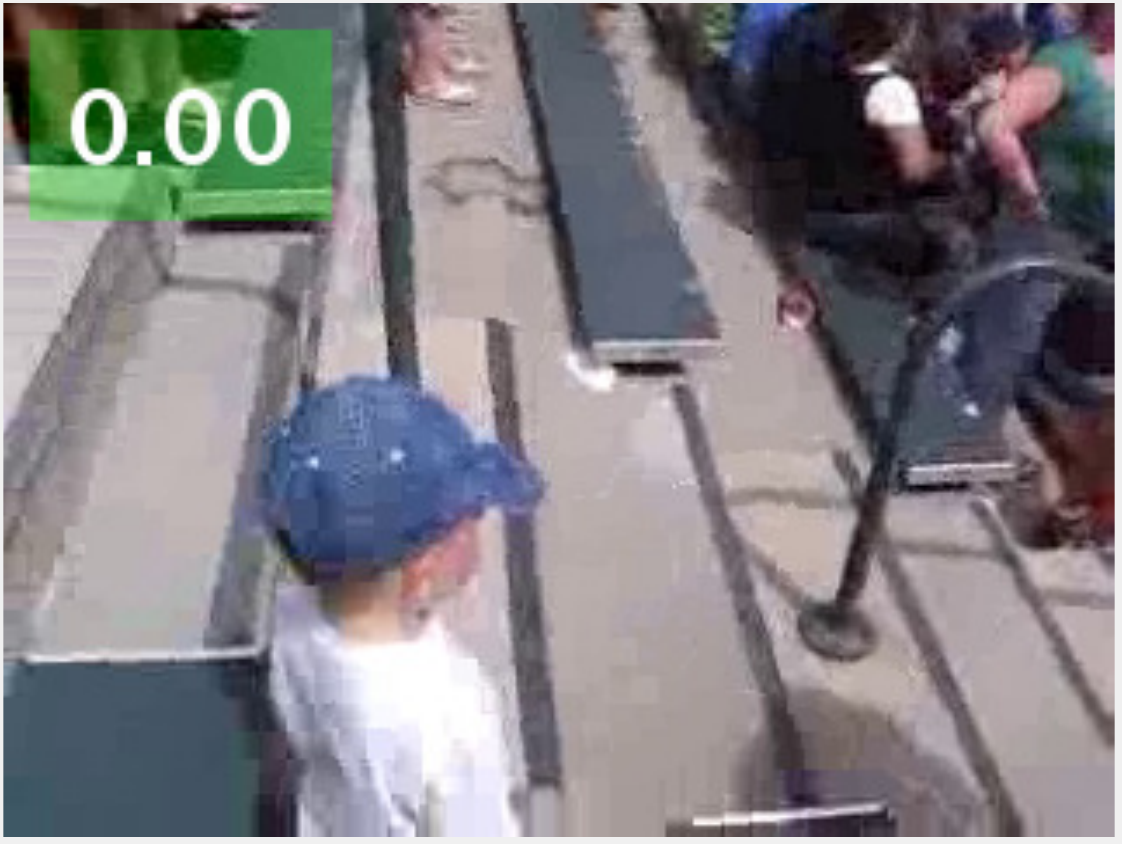} 
   \includegraphics[width=0.093\linewidth]{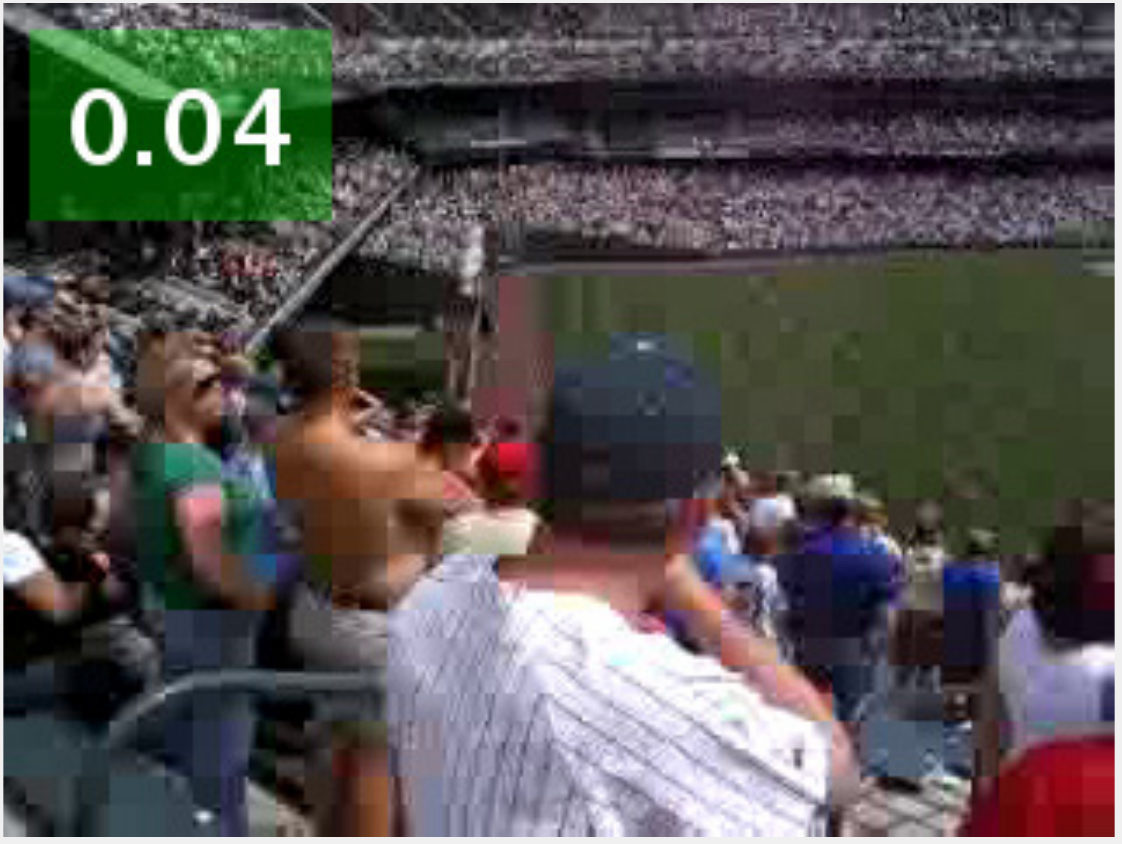}\\
      \small{Event: baseball}
\end{tabular}
\end{center}
\vspace{-2mm}
   \caption{ The calculated attention weights of our TAGM model for examples from test set of CCV database.  The attention weight is indicated for representative frames. Our TAGM is able to capture the action of `riding bike' for the event `biking', `cake' for the event `birthday' and `infield zone' for `baseball'. A video containing these three complete sample sequences  is presented in the supplementary material.}
\label{fig:CCV_correct}
\end{figure*}
\vspace{-5mm}
\subsubsection{Experimental Setup}
\vspace{-2mm}
Following Jiang et al~\cite{CCV}, we use the same split for training and test sets: 4659 videos as the training set and 4658 as the test set. We compare our model with the baseline method~\cite{CCV_superfastsvm} on this dataset, which performs classification separately with Support Vector Machine (SVM) models trained on the  bag-of-words representations for several popular features separately and then combines the results  using late fusion. Its experimental results show that Convolutional Neural networks (CNNs) features perform best among all features they tried, hence we choose to use CNN features with the same setup, i.e., the outputs (4,096 dimensions) of the seventh fully-connected layer of a pre-trained AlexNet model~\cite{AlexNet}. For the sake of computational efficiency,  we extract CNN features with a sampling rate 1/8 (one out of every eight frame).\\
We adopt mean Average Precision (mAP) as the evaluation metric, which is typically used for CCV dataset \cite{CCV, CCV_superfastsvm}. Since more than one event (correct label) can happen in a sample, we perform binary classification for each category but train them jointly, hence the prediction score for each category is calculated by a sigmoid function instead of softmax Equation~\ref{eqn:softmax}:
\begin{small}
{\setlength{\abovedisplayskip}{6pt}
\setlength{\belowdisplayskip}{6pt}
\begin{align}
P(y_k=1 | \mathbf{h}_{T}) = \frac{1}{1+\mathrm{exp}\{-(\mathbf{W}^\top_k \mathbf{h}_T + b_k)\}}
\label{eqn:ccv_sig}
\end{align}}
\end{small}\\
and joint binary cross-entropy over $K$ categories is minimized: 
\begin{small}
{\setlength{\abovedisplayskip}{6pt}
\setlength{\belowdisplayskip}{6pt}
\begin{align*}
\mathcal{L}\! = \! -\sum_{n=1}^N \sum_{k=1}^K \Big [  \log P(y_k=1 | \mathbf{h}_{T}) + \log(1-P(y_k=0 |  \mathbf{h}_{T}))  \Big ]
\label{eqn:ccv_loss}
\end{align*}}
\vspace{-7mm}
\end{small}
\subsubsection{Results and Discussion}
\vspace{-2mm}
\paragraph{Evaluation of Classification Performance.}
We compare our model with the event recognition system proposed by dataset authors~\cite{CCV_superfastsvm}. Table~\ref{table:CCV} presents the performance of several models for event recognition, in which our TAGM outperforms the other recurrent models by a large margin.  The baseline BOW+SVM employs the one-vs-all strategy to train a separate classifier for each event while our model trains all events jointly in a single classifier. Our model still shows encouraging results since it is quite a challenging task for TAGM to capture salient sections for 20 events with complex scenes simultaneously.  Moreover, our TAGM can provide a meaningful interpretation which the baseline models cannot do.
\vspace{-2.5mm}
\begin{table}[!htb]
\vspace{-2mm}
\caption{Mean Average Precision (mAP) of our TAGM model and baseline models on CCV dataset. }
\centering
\vspace{2mm}
\normalsize
\renewcommand\arraystretch{1.2}
\resizebox{0.95\linewidth}{!}{
\begin{tabular}{ l | l | l | c}
\Xhline{1pt}
\hline
Model & Training strategy & Feature & mAP \\
\hline
\multirow{4}{*}{\tabincell{l}{BOW+SVM\\+late average fusion}} &\multirow{4}{*}{ \tabincell{l}{Separately\\ (one-vs-all)  }}& SIFT & 0.52 \\
& & STIP & 0.45 \\
& & SIFT+STIP & 0.55 \\
& & CNN & 0.67 \\
\hline
Plain-RNN & Jointly & CNN & 0.45\\
GRU & Jointly & CNN & 0.56\\
LSTM & Jointly & CNN & 0.55\\
\hline
\multirow{1}{*}{TAGM }  & Jointly &CNN& 0.63 \\
\hline
\end{tabular}}
\label{table:CCV}
\vspace{-3mm}
\end{table}
\vspace{-3mm}
\paragraph{Sequence Salience Detection.}
Salience detection for CCV database is a difficult but appealing task due to complex and long scenes in videos. Figure~\ref{fig:CCV_correct} shows some examples where TAGM correctly locates the salient subsequences by the attention weights.  Our model is able to capture the relevant action, object and scene to the event, e.g., the action of riding bike for the event `biking', cake for the event `birthday' and baseball playground for the event `baseball'. It is interesting to note that the frame with the score 0.42 in event `baseball' achieves the high score probably because of the real-time screen in the top right corner.
\vspace{-6mm}

\section{Conclusion}
\vspace{-2mm}
In this work, we presented the Temporal Attention-Gated Model (TAGM), a new model for classifying noisy and unsegmented sequences. The model is inspired by attention models and gated recurrent networks and is able to detect salient parts of the sequence while ignoring irrelevant and noisy ones. The resulting hidden representation suffers less from the effect of noise and and thus leads to better performance.  Furthermore, the learned attention scores provide a physically meaningful interpretation of relevance of each time step observation for the final decision. We showed the generalization of our approach on three very different datasets and sequence classification tasks. As future work, our model could be extended to help with document or video summarization.


{\small
\bibliographystyle{ieee}
\bibliography{egbib}
}

\end{document}